\pdfoutput=1
\documentclass[11pt]{article}

\usepackage{EMNLP2022}

\usepackage{times}
\usepackage{latexsym}
\usepackage{graphicx}
\usepackage{subfig}
\usepackage{multirow}
\usepackage{array}
\usepackage{booktabs}
\usepackage{xcolor}
\usepackage{amsmath,amssymb}
\usepackage[geometry]{ifsym}
\usepackage{color}
\usepackage{tikz}
\usetikzlibrary{shapes}

\newcolumntype{P}[1]{>{\centering\arraybackslash}p{#1}}

\usepackage[T1]{fontenc}
\usepackage[utf8]{inputenc}

\usepackage{microtype}

\usepackage{inconsolata}
\usepackage{upquote}

\usepackage{threeparttable}

\title{When Can Transformers Ground and Compose: Insights from Compositional Generalization Benchmarks}

\usepackage{pifont}

\author{\stepcounter{footnote}Ankur Sikarwar\thanks{\enspace Work done at Microsoft Research India.}$\ \,^1$ \quad Arkil Patel\footnotemark[2]$\ \,^{2,3}$ \quad Navin Goyal$\,^4$\\
	$^1$I2R, A*STAR Singapore \quad
	$^2$Mila - Quebec AI Institute \quad
	$^3$McGill University \\
	$^4$Microsoft Research India
	\\
	{\tt ankursikarwar.as@gmail.com} \\
	{\tt arkil.patel@gmail.com, navingo@microsoft.com} \\
}

\begin{document}
\maketitle
\begin{abstract}

Humans can reason compositionally whilst grounding language utterances to the real world. Recent benchmarks like ReaSCAN \cite{reascan} use navigation tasks grounded in a grid world to assess whether neural models exhibit similar capabilities. In this work, we present a simple transformer-based model that outperforms specialized architectures on ReaSCAN and a modified version \cite{google_multimodal} of gSCAN \cite{gscan}. On analyzing the task, we find that identifying the target location in the grid world is the main challenge for the models. Furthermore, we show that a particular split in ReaSCAN, which tests depth generalization, is unfair. On an amended version of this split, we show that transformers can generalize to deeper input structures. Finally, we design a simpler grounded compositional generalization task, \textbf{RefEx}, to investigate how transformers reason compositionally. We show that a single self-attention layer with a single head generalizes to novel combinations of object attributes. Moreover, we derive a precise mathematical construction of the transformer’s computations from the learned network. Overall, we provide valuable insights about the grounded compositional generalization task and the behaviour of transformers on it, which would be useful for researchers working in this area.

\end{abstract}

\section{Introduction}

Natural Languages are believed to be compositional \cite{partee_compositionality}, i.e., the meaning of an expression is determined by the meaning of its constituents and how they are combined. The field of \emph{compositional generalization} seeks to understand whether neural models used for language processing exhibit compositional behaviour. In recent years, the field has received increased attention resulting in the development of many new benchmarks \cite{scan, cogs, cfq} and approaches \cite{prim_subs, lake_meta, ness, lear} to solve them.

\begin{figure}[t]
	\centering
	\includegraphics[scale=0.45]{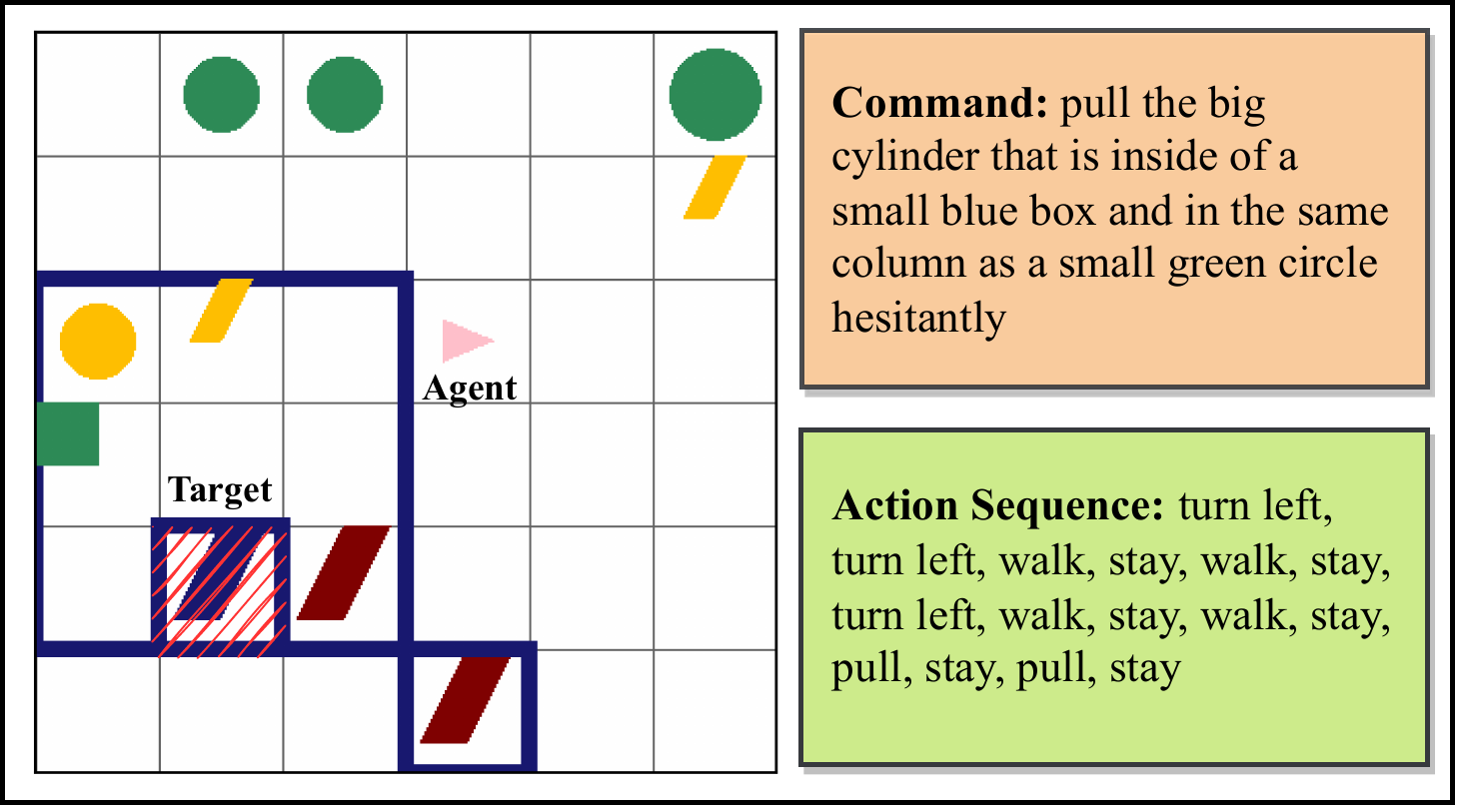}
	\caption{\label{fig:reascan_example} An example from the ReaSCAN dataset.}
\end{figure}

Natural language utterances are also grounded to the real world. To encourage development of systems that are both \emph{compositional} and \emph{grounded}, \citet{gscan} created the gSCAN dataset. Recently, \citet{google_multimodal} proposed the GSRR dataset\footnote{They proposed new test splits for gSCAN, which we call Grounded Spatial Relation Reasoning (GSRR).} and \citet{reascan} proposed the ReaSCAN dataset to address certain limitations in gSCAN. These tasks consist of navigation commands grounded in a 2D grid world containing an agent and multiple objects with different visual attributes. Given a command and grid world, a model needs to output the sequence of actions for the agent to execute. Fig. \ref{fig:reascan_example} shows an example from ReaSCAN. The difficulty of the task lies in generalizing to out-of-distribution splits that are formed by systematically holding out particular compositions of object attributes and command structures from train set. \citet{heinze, kuo} developed specialized architectures for gSCAN that are either difficult to adapt to other problems or require extra supervision.

\textbf{Contributions.} Our goal is to better understand these grounded compositional generalization tasks and design \emph{generic} ML models to solve them. Our contributions include:

\textbf{(i)} We propose the \textbf{Gro}unded \textbf{Co}mpositional \textbf{T}ransformer (GroCoT), which was created by making simple and well-motivated modifications to a multi-modal transformer model \cite{google_multimodal}. GroCoT achieves state-of-the-art performances on both, GSRR and ReaSCAN.\footnote{We make our source code and data available at: \href{https://github.com/ankursikarwar/Grounded-Compositional-Generalization}{https://github.com/ankursikarwar/Grounded-Compositional-Generalization}.} Our results clearly show that simple transformer-based models generalize well on these tasks. 

\textbf{(ii)} We design a series of experiments to understand the underlying challenges in these tasks. We show that identifying the target location, rather than sequence generation, is the main difficulty. We also demonstrate that the split, testing depth generalization in ReaSCAN is unfair in that the training data does not provide the models with sufficient information to correctly choose among competing hypotheses. On experimenting with a modified training distribution, we show that simple transformer-based models can successfully generalize to commands with greater depths.

\textbf{(iii)} We examine why transformers are so successful at generalizing compositionally on these tasks. To this end, we introduce a new task called \textbf{RefEx} (`Referring Expressions'), which provides a simpler setting isolating some of the main features of ReaSCAN. We find that a 1-layer, 1-head attention-only transformer is capable of grounding and generalizing to novel compositions of multiple visual attributes; moreover, it admits a complete interpretation of the computations. RefEx also allows easier probing and leads us to identify and solve an overfitting issue with transformers on a particular ReaSCAN split.

\section{Background}

\begin{table}[t]
	\small{\centering
		\begin{tabular}{m{22em}}
			\toprule
			\texttt{Simple:} \\
			\midrule
			\footnotesize{Walk to the small red square.}\\
			\midrule

			\texttt{1-relative-clause:} \\
			\midrule
			\footnotesize{Pull the blue circle \textbf{that is} in the same row as the small green square.}\\
			\midrule

			\texttt{2-relative-clause:} \\
			\midrule
			\footnotesize{Push the small blue cylinder \textbf{that is} in the same column as the big green circle \textbf{and} the red square.}\\

			\bottomrule
		\end{tabular}
		\caption{\label{tab:command_ex} Different types of ReaSCAN commands.}
	}
\end{table}

We focus on the gSCAN \cite{gscan}, GSRR \cite{google_multimodal} and ReaSCAN \cite{reascan} datasets. A model, provided with a natural language command, is tasked with generating a sequence of actions to navigate an agent in a 2D grid world populated with objects. Below, we shall explain the setting of the ReaSCAN task in detail. More information about other datasets is provided in Appendix \ref{app:datasets}.

Each example consists of a $d \times d$ grid world ($d=6$), a natural language command and the corresponding output sequence. Each cell in the grid world is described by a $c$-dimensional vector that concatenates one-hot encodings for the three object attributes, color $\mathcal{C} = \{\mathsf{red}, \mathsf{green}, \mathsf{blue}, \mathsf{yellow}\}$, shape $\mathcal{S} = \{\mathsf{circle}, \mathsf{square}, \mathsf{cylinder}, \mathsf{box}\}$, and size $\mathcal{D} = \{1,2,3,4\}$ along with information about agent orientation $\mathcal{O} = \{\mathsf{left}, \mathsf{right}, \mathsf{up}, \mathsf{down}\}$ and agent presence $\mathcal{B} = \{\mathsf{yes}/ \mathsf{no}\}$. Hence, the entire grid world is represented as a tensor $\mathbf{W} \in \mathbb{R}^{d\times d \times c}$. The natural language command $x := (x_1, x_2, \ldots, x_n)$ is generated using a context-free grammar (CFG), which is described in Appendix \ref{app:reascan}. ReaSCAN has three types of input commands which we illustrate in Table \ref{tab:command_ex}. The output sequence $y := (y_1, y_2, \ldots, y_m)$ is made up of a finite set of action tokens $\mathcal{A} = \{\mathsf{walk}, \mathsf{push}, \mathsf{pull}, \mathsf{stay}, \mathsf{turn\ left}, \mathsf{turn\ right}\}$.

The main challenge of the task is generalizing on the specially designed test splits that consist of various types of examples systematically held-out from the train set as shown in Table \ref{tab:reascan_splits} (more details in Appendix \ref{app:reascan}).

The results of various previously proposed methods are shown in Table \ref{tab:google_result} for GSRR, Table \ref{tab:main_result} for ReaSCAN, and Table \ref{tab:gscan_result} for gSCAN. \citet{google_multimodal} outperformed all previous methods \cite{gao, kuo, heinze} on gSCAN and GSRR. Hence, for ReaSCAN, we don't re-implement those methods as baselines; rather, we compare directly against \citet{google_multimodal}.

\begin{table}[t]
	\small{\centering
		\begin{tabular}{P{3em}p{18em}}
			\toprule
			\textsc{Split} & \textsc{Held-out Examples} \\
			\midrule
			A1 & \emph{yellow squares} referred with color and shape \\
			A2 & \emph{red squares} as target \\
			A3 & \emph{small cylinders} referred with size and shape \\
			B1 & \emph{small red circle} and \emph{big blue square} co-occur \\
			B2 & \emph{same size as} and \emph{inside of} relations co-occur \\
			C1 & additional conjunction clause added to \texttt{2-relative-clause} commands \\
			C2 & \texttt{2-relative-clause} command with \emph{that is} instead of \emph{and} \\
			\bottomrule
		\end{tabular}
		\caption{\label{tab:reascan_splits} Compositional splits in ReaSCAN.}
	}
\end{table}

\begin{table*}[t]
	\small{\centering
		\begin{tabular}{p{15.1em}P{5.5em}P{7.5em}P{2em}P{2em}P{2em}P{2em}P{2em}}
			\toprule
			\small{\textsc{Model}} & \small{\textsc{Random (\uppercase\expandafter{\romannumeral1\relax})}} & 
			\small{\textsc{Comp. Average}} &
			\small{\textsc{\uppercase\expandafter{\romannumeral2\relax}}} &
			\small{\textsc{\uppercase\expandafter{\romannumeral3\relax}}} &
			\small{\textsc{\uppercase\expandafter{\romannumeral4\relax}}} &
			\small{\textsc{\uppercase\expandafter{\romannumeral5\relax}}} &
			\small{\textsc{\uppercase\expandafter{\romannumeral6\relax}}}
			\\
			
			\midrule
			
			Multimodal LSTM 
			\scriptsize{\cite{reascan}} &
			86.5 &
			58.9 &
			40.1 &
			86.1 & 
			5.5 & 
			81.4 &
			81.8 
			\\
			
			Multimodal Transformer \scriptsize{\cite{google_multimodal}} &
			94.7 &
			63.5 &
			64.4 &
			94.9 &
			49.6 & 
			59.3 & 
			49.5 
			\\
			
			GroCoT (ours) &
			\textbf{99.9} &
			\textbf{98.8} &
			\textbf{98.6} &
			\textbf{99.9} &
			\textbf{99.7} & 
			\textbf{99.5} & 
			\textbf{96.5}
			\\
			
			\bottomrule
		\end{tabular}
		\caption{\label{tab:google_result} Performance of GroCoT on GSRR \cite{google_multimodal} in comparison to baselines and previous approaches.}
	}
\end{table*}

\begin{table*}[t]
	\small{\centering
		\begin{tabular}{p{15.1em}P{5em}P{2em}P{2em}P{2em}P{2em}P{2em}P{2em}P{2em}}
			\toprule
			\small{\textsc{Model}} & \small{\textsc{Average}} & \small{\textsc{A1}} & 
			\small{\textsc{A2}} &
			\small{\textsc{A3}} &
			\small{\textsc{B1}} &
			\small{\textsc{B2}} &
			\small{\textsc{C1}} &
			\small{\textsc{C2}}
			\\
			
			\midrule
			
			Multimodal LSTM
			\scriptsize{\cite{reascan}} &
			40.4 &
			50.4 &
			14.7 &
			50.9 & 
			52.2 & 
			39.4 &
			49.7 &
			25.7
			\\
			
			GCN-LSTM
			\scriptsize{\cite{gao}} &
			60.5 &
			92.3 &
			42.1 &
			87.5 & 
			69.7 & 
			52.8 &
			57.0 &
			22.1
			\\
			
			Multimodal Transformer \scriptsize{\cite{google_multimodal}} &
			69.9 &
			96.7 &
			58.9 &
			93.3 & 
			79.8 & 
			59.3 &
			75.9 &
			25.5
			\\
			
			GroCoT w/ vanilla self-attention &
			80.8 &
			99.2 &
			88.1 &
			98.7 & 
			\textbf{94.6} & 
			\textbf{86.4} &
			75.3 &
			23.4
			\\
			
			GroCoT (ours) &
			\textbf{82.2} &
			\textbf{99.6} &
			\textbf{93.1} &
			\textbf{98.9} & 
			93.9 & 
			86.0 &
			\textbf{76.3} &
			\textbf{27.3}
			\\
			
			\bottomrule
		\end{tabular}
		\caption{\label{tab:main_result} Performance of GroCoT on ReaSCAN \cite{reascan} in comparison to baselines and previous approaches.}
	}
\end{table*}

\begin{table*}[t]
	\small{\centering
		\begin{tabular}{P{2.5em}P{2.5em}P{2.5em}P{5em}P{2.5em}P{2.5em}P{2.5em}P{2.5em}P{2.5em}P{2.5em}P{2.5em}}
			\toprule
			\small{\textsc{ISR}} & 
			\small{\textsc{ISA}} & 
			\small{\textsc{EM}} & 
			\small{\textsc{Average}} & 
			\small{\textsc{A1}} & 
			\small{\textsc{A2}} &
			\small{\textsc{A3}} &
			\small{\textsc{B1}} &
			\small{\textsc{B2}} &
			\small{\textsc{C1}} &
			\small{\textsc{C2}}
			\\
			
			\midrule
			
			\ding{55} &
			\ding{55} &
			\ding{55} &
			69.9 &
			96.7 &
			58.9 &
			93.3 & 
			79.8 & 
			59.3 &
			75.9 &
			25.5
			\\
			
			\ding{51} &
			\ding{55} &
			\ding{55} &
			70.7 &
			96.4 &
			75.2 &
			93.8 & 
			78.2 & 
			57.9 &
			71.7 &
			21.9
			\\
			
			\ding{55} &
			\ding{51} &
			\ding{55} &
			77.7 &
			99.1 &
			87.6 &
			98.4 & 
			89.2 & 
			67.7 &
			\textbf{79.3} &
			22.3
			\\
			
			\ding{55} &
			\ding{55} &
			\ding{51} &
			73.6 &
			97.4 &
			70.7 &
			94.3 & 
			83.2 & 
			66.4 &
			76.7 &
			26.3
			\\
			
			\ding{51} &
			\ding{51} &
			\ding{55} &
			79.0 &
			99.0 &
			90.9 &
			98.2 & 
			88.3 & 
			72.7 &
			77.0 &
			26.9
			\\
			
			\ding{55} &
			\ding{51} &
			\ding{51} &
			77.8 &
			98.5 &
			79.6 &
			97.8 & 
			90.8 & 
			78.9 &
			78.6 &
			20.8
			\\

			\ding{51} &
			\ding{55} &
			\ding{51} &
			73.9 &
			97.7 &
			76.7 &
			96.0 & 
			82.1 & 
			64.8 &
			75.1 &
			25.1
			\\
			
			\ding{51} &
			\ding{51} &
			\ding{51} &
			\textbf{82.2} &
			\textbf{99.6} &
			\textbf{93.1} &
			\textbf{98.9} & 
			\textbf{93.9} & 
			\textbf{86.0} &
			76.3 &
			\textbf{27.3}
			\\
			
			\bottomrule

		\end{tabular}
		\caption{\label{tab:ablations_reascan} Ablation study for GroCoT on ReaSCAN \cite{reascan}. The results show that all our modifications are necessary to achieve best performance. ISR, ISA, and EM stand for Improved Spatial Representation, Interleaving Self-Attention, and Embedding Modification, respectively.}
	}
\end{table*}

\section{Our Approach}

We start with the multimodal transformer model as used in \citet{google_multimodal}. This model, hereafter called the \emph{base model}, follows encoder-decoder structure \citet{transformer} and uses cross-modal attention in the encoder.

\textbf{Encoder} maps world state $\mathbf{W} \in \mathbb{R}^{d\times d \times c}$ to visual representation $\mathbf{H}^v$ through multi-scale CNNs followed by linear layers. The command tokens $x := (x_1, x_2, \ldots, x_n)$ are encoded into embeddings $\mathbf{H}^l = \{\mathbf{h}^l_1, \mathbf{h}^l_2, \ldots, \mathbf{h}^l_n\}$. These are passed through $N$ transformer blocks, each consisting of two parallel multi-head attention blocks (one for vision and one for language modality), with representation of one modality passed as key and value to the attention block of the other modality.

\textbf{Decoder} consists of $N$
stacked blocks similar to the decoder in \citet{transformer}. Each block contains one self-attention block and one multi-head attention block over the contextual representation $\mathbf{H_\mathit{c}} = [\mathbf{H_\mathit{l}};  \mathbf{H_\mathit{v}}]$ of the encoder.

Below, we describe the modifications we make to this base architecture to create GroCoT. Implementation details are provided in Appendix \ref{app:implement}.

\textbf{Improving Spatial Representation.} For this task, models need to perform spatial reasoning between objects that may possibly be very far from each other in the grid world. The base model \cite{google_multimodal} employed a multi-scale CNN to encode the world state $\mathbf{W}$ before feeding it to the transformer. However, CNNs, without the presence of large filters (i.e., large receptive fields), are inept at understanding the spatial relationships between parts of the image that are not in immediate vicinity. To address this limitation, instead of passing the world state tensor $\mathbf{W}$ through a multi-scale CNN, we propose tokenizing the grid cells and projecting them to a higher dimension  $\mathbf{W}^v = \{\mathbf{w}^v_1, \mathbf{w}^v_2, \ldots, \mathbf{w}^v_{d\times d}\}, \mathbf{w}^v_i \in \mathbb{R}^{d_\mathrm{model}}$. In line with \citet{vilbert}, we separately encode the spatial information of grid cells in a 2D vector, where the first dimension holds the row value and the second one holds the column value. We project these spatial encodings to a higher dimension $\mathbf{S}^v = \{\mathbf{s}^v_1, \mathbf{s}^v_2, \ldots, \mathbf{s}^v_{d\times d}\}, \mathbf{s}^v_i \in \mathbb{R}^{d_\mathrm{model}}$ and add them to their corresponding grid cell representations to obtain the final grid cell embedding input to the transformer $\mathbf{H}^v = \{\mathbf{h}^v_1, \mathbf{h}^v_2, \ldots, \mathbf{h}^v_{d\times d}\}, \mathbf{h}^v_i = \mathbf{w}^v_i + \mathbf{s}^v_i \in \mathbb{R}^{d_\mathrm{model}}$.

\textbf{Interleaving Self-Attention.} The base model uses cross-modal attention in all encoder layers. While this facilitates grounding of semantic information across both modalities, we believe this method to be inefficient. We know that different layers in both vision transformers and language transformers encode different levels of semantic knowledge \cite{vit, do_vision, bertdoes}. To allow efficient grounding, we want both visual and language modality streams to develop their own representations before synchronizing them with each other via cross-modal attention. Hence, we propose interleaving self-attention layers between co-attention layers to allow intra-modal interaction within each stream before cross-modal interaction.

\begin{figure}[t]
	\centering
	\includegraphics[scale=0.4]{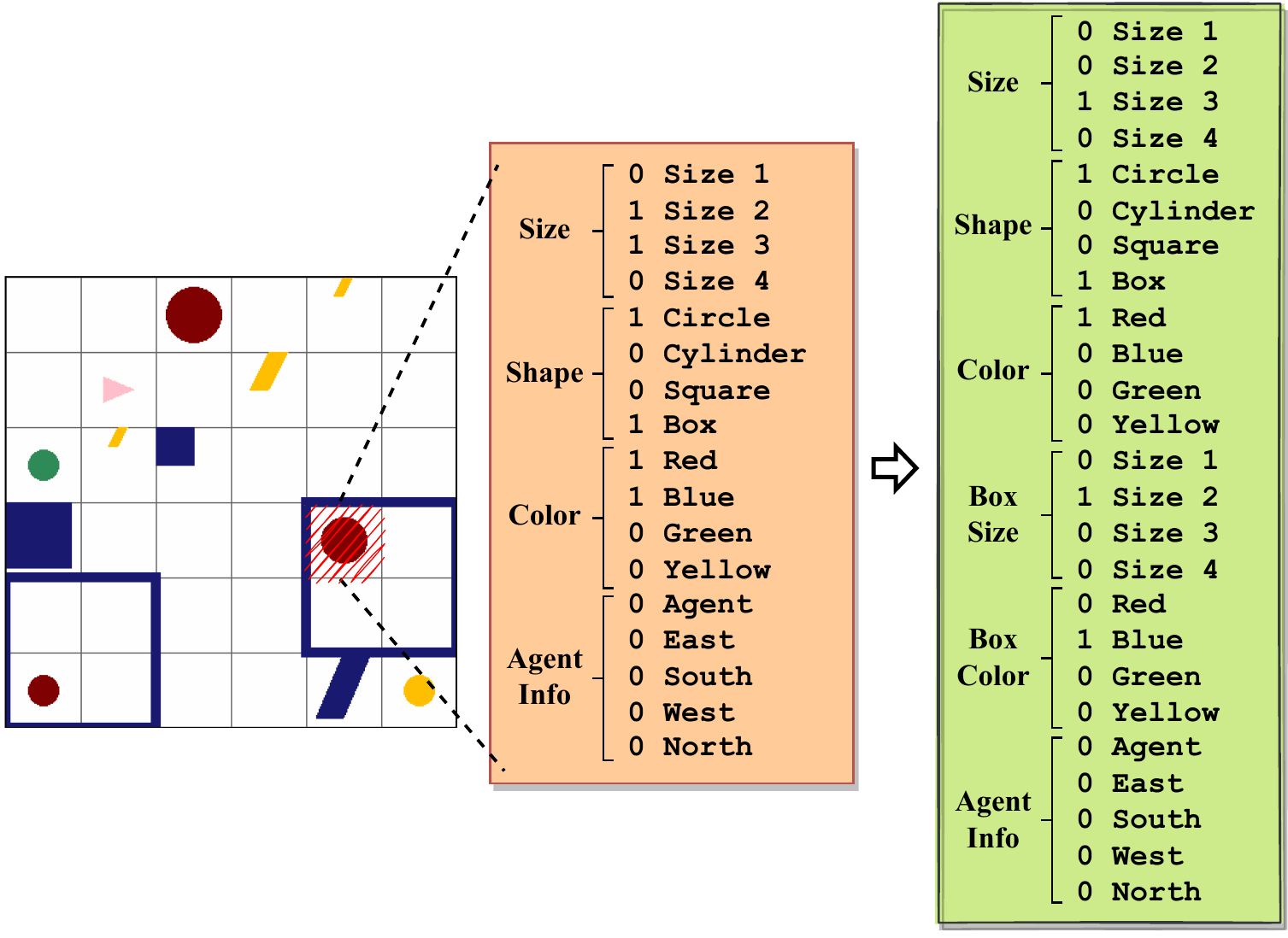}
	\caption{\label{fig:mod_world_state} Modified Grid Cell Encoding}
\end{figure}

\textbf{Modified World State Encoding.} In ReaSCAN, for particular examples where another object is present in the top left corner of a box object, the grid cell embedding corresponding to that corner is calculated by adding up the vector encodings corresponding to the object and the box \cite{reascan}. However, this design is inherently flawed because the attributes of the two objects cannot be disambiguated from the sum of their individual encodings. This issue causes models to fail in such examples. To handle such cases, we propose using a higher dimensional grid cell embedding (see Fig. \ref{fig:mod_world_state}) to represent color and size properties of the box separately from other objects.

\textbf{Discussion.} The results are provided in Table \ref{tab:google_result} for GSRR and Table \ref{tab:main_result} for ReaSCAN. We also provide exhaustive ablations of our approach on ReaSCAN in Table \ref{tab:ablations_reascan}. On both datasets, GroCoT outperforms all specialized architectures. From the ablation study on ReaSCAN, we observe that both, improved spatial representation and interleaved self-attention, lead to significant improvements. The modified embedding structure additionally helps when examples contain box objects (see improvement in ReaSCAN B2 split). Our model also saturates performance on most splits in gSCAN (see Table \ref{tab:gscan_result}). We also evaluated the effect of using vanilla self-attention (as used in the original Transformer \cite{transformer}) in GroCoT\footnote{Note that our other proposed modifications (improved spatial representation and embedding modification) are still being applied.} on ReaSCAN and found that it achieves surprisingly high accuracies (see Table \ref{tab:main_result}). Our hypothesis is that vanilla self-attention facilitates individual processing of both modalities similar to our interleaving self-attention approach, and hence it does not hurt the model performance significantly.

Overall, our results show that \emph{a simple transformer-based model is capable of generalizing compositionally on most of the challenges proposed by gSCAN, GSRR and ReaSCAN.} Instead of presenting GroCoT as a broadly applicable method solving compositional generalization, we only wish to establish that such simple transformer-based models can exhibit strong compositional generalization capabilities and serve as powerful baselines.

\section{Analyzing the Grounded Compositional Generalization Tasks}\label{sec:analysis}

\subsection{Target Identification vs Navigation: What is the Challenge?}\label{sec:target_loc}

\begin{table*}[th]
	\footnotesize{\centering
		\begin{tabular}{p{17.5em}P{4em}P{2.3em}P{2.3em}P{2.3em}P{2.3em}P{2.3em}P{2.3em}P{2.3em}}
			\toprule
			\small{\textsc{}} & \small{\textsc{Average}} & \small{\textsc{A1}} & 
			\small{\textsc{A2}} &
			\small{\textsc{A3}} &
			\small{\textsc{B1}} &
			\small{\textsc{B2}} &
			\small{\textsc{C1}} &
			\small{\textsc{C2}}
			\\
			
			\midrule
			
			Target Identification accuracy &
			78.5 &
			96.4 &
			85.7 &
			95.4 &
			90.4 & 
			83.5 & 
			70.2 &
			27.7
			\\
			
			Error overlap with ReaSCAN model &
			88.3 &
			89.5 &
			87.9 &
			83.5 & 
			87.4 & 
			86.7 &
			88.5 &
			94.7
			\\
			
			\midrule
			
			ReaSCAN accuracy w/ gold target locations &
			99.9 &
			99.9 &
			99.9 &
			99.9 & 
			99.9 & 
			99.9 &
			99.9 &
			99.9
			\\
			
			\bottomrule
		\end{tabular}
		\caption{\label{tab:target_loc} Experimental results for GroCoT to understand the performance bottleneck in ReaSCAN.}
	}
\end{table*}

\begin{table*}[t]
	\small{\centering
		\begin{tabular}{P{2.5em}P{2.5em}P{2.5em}P{5em}P{2.5em}P{2.5em}P{2.5em}P{2.5em}P{2.5em}P{2.5em}P{2.5em}}
			\toprule
			\small{\textsc{ISR}} & 
			\small{\textsc{ISA}} & 
			\small{\textsc{EM}} & 
			\small{\textsc{Average}} & 
			\small{\textsc{A1}} & 
			\small{\textsc{A2}} &
			\small{\textsc{A3}} &
			\small{\textsc{B1}} &
			\small{\textsc{B2}} &
			\small{\textsc{C1}} &
			\small{\textsc{C2}}
			\\
			
			\midrule
			
			\ding{55} &
			\ding{55} &
			\ding{55} &
			69.8 &
			94.3 &
			59.4 &
			91.0 & 
			79.9 & 
			64.3 &
			74.4 &
			25.1
			\\
			
			\ding{51} &
			\ding{55} &
			\ding{55} &
			71.9 &
			95.4 &
			87.0 &
			94.1 & 
			75.7 & 
			56.0 &
			71.6 &
			23.5
			\\
			
			\ding{55} &
			\ding{51} &
			\ding{55} &
			77.7 &
			96.0 &
			84.3 &
			95.5 & 
			90.5 & 
			73.4 &
			78.2 &
			25.7
			\\
			
			\ding{55} &
			\ding{55} &
			\ding{51} &
			77.0 &
			95.5 &
			76.3 &
			93.4 & 
			87.8 & 
			81.2 &
			75.8 &
			\textbf{29.1}
			\\
			
			\ding{51} &
			\ding{51} &
			\ding{55} &
			74.6 &
			95.6 &
			85.4 &
			95.1 & 
			84.7 & 
			69.8 &
			71.4 &
			20.0
			\\
			
			\ding{55} &
			\ding{51} &
			\ding{51} &
			\textbf{81.1} &
			\textbf{96.5} &
			\textbf{89.5} &
			\textbf{96.5} & 
			\textbf{91.9} & 
			\textbf{88.1} &
			\textbf{80.7} &
			24.3
			\\

			\ding{51} &
			\ding{55} &
			\ding{51} &
			72.7 &
			95.7 &
			71.7 &
			93.7 & 
			82.0 & 
			73.2 &
			71.4 &
			21.2
			\\
			
			\ding{51} &
			\ding{51} &
			\ding{51} &
			78.5 &
			96.4 &
			85.7 &
			95.4 & 
			90.4 & 
			83.5 &
			70.2 &
			27.7
			\\
			
			\bottomrule

		\end{tabular}
		\caption{\label{tab:ablations_target} Target Identification accuracy for different ablations on GroCoT when tested on ReaSCAN \cite{reascan}. ISR, ISA, and EM stand for Improved Spatial Representation, Interleaving Self-Attention, and Embedding Modification, respectively.}
		
	}
\end{table*}

In order to solve these tasks, a model needs to perform two subtasks: (1) identify the target location by composing the words and reasoning about the relative clauses, and (2) navigate the agent in the grid world by generating the right set of output tokens. To understand why models fail and how to improve them, we need to pinpoint where the main difficulty in the task lies. Below we describe a set of experiments that demonstrate that target identification is the main challenge in ReaSCAN rather than navigation or sequence generation.\footnote{Studies with similar objectives have also been carried out by past work. We explain their limitations and compare against them in Appendix \ref{app:target_loc_prev_work}.} We show this for the gSCAN dataset in Appendix \ref{app:target_loc_gscan}.

\textbf{Target Identification from Encoder Representations.} We train a linear layer on top of the last layer of learned encoder representations of the best-performing model to perform target identification.\footnote{We also experimented with predicting the target location from earlier encoder representations and random vectors to serve as baselines. These results are provided in Figure \ref{fig:linear_probe}.} We model the task as a 36-way classification problem, where each grid location is treated as a distinct class. We train the model over all ReaSCAN examples with the ground truth target locations. Note that we only update the weights of the linear layer; the parameters of the encoder are kept frozen. We then test this model's target identification abilities over the systematic generalization splits.

The first row of Table \ref{tab:target_loc} shows the performance of the model on target identification. We see the same trend in performance across all splits as we saw for the full model on ReaSCAN (see Table \ref{tab:main_result}). This indicates that target identification might be the main difficulty for the model. To illustrate more concretely, we calculate the overlap of errors between the target identification model and the ReaSCAN model. As can be seen from the second row in Table \ref{tab:target_loc}, for each split, out of all the examples where the ReaSCAN model failed, the percentage of examples where the target identification model also failed is extremely high. 

We also provide exhaustive ablations for this experiment in Table \ref{tab:ablations_target}. Our proposed modifications do indeed enable better target identification. However, there might be other minor aspects of the problem that are tackled by our modifications (model with ISR performs better overall on ReaSCAN but is not the best on target identification). Also note that these experiments are performed without the decoder, which is essential in solving ReaSCAN.

\textbf{Sequence Generation from Gold Target Locations.} We provide the model with gold target locations when training it on the ReaSCAN training set. We enumerate all the 36 grid cell locations and simply append $\mathsf{gridnum}$ to the end of the natural language command where $\mathsf{gridnum} \in \{1, 2, ..., 36\}$, depending on the ground-truth target location for a particular example. The results are provided in the last row of Table \ref{tab:target_loc}. Clearly, the model is able to generalize almost perfectly when provided with the ground-truth target locations. This shows that if the target is identified, the model has no difficulty in navigating the agent towards it.

In this section, with comprehensive empirical evidence, we showed that \emph{models are highly competent at agent navigation} and that \emph{the chief difficulty lies in identifying the target location.}

\subsection{Issues in ReaSCAN Test Set Design}\label{sec:issue_c2}

Compositional generalization setups are used to assess specific capabilities of models. However, if the train-test splits within these setups are not carefully created, then the experimental results may lead to false conclusions \cite{revisiting}. In this section, we show that the C2 test set of ReaSCAN is unfair because of lack of necessary information in the train set. We then propose a correction in the train-test setup that allows us to fairly evaluate the depth generalization capabilities of models.

\textbf{C2 Test Set is Unfair.} The train set of ReaSCAN consists of commands with different structures as shown in Table \ref{tab:command_ex}. The C2 test set is made up of commands with the other type of \texttt{2-relative-clause} structure (e.g., ``walk to the red square \emph{that is} in the same row as the blue cylinder \emph{that is} in the same column as a green circle''). This split tests whether a model is able to perform recursion to higher depths. It is clear that the only difference between the \texttt{2-relative-clause} commands in the train set and the C2 test set is that the `and' connecting the last two clauses is replaced with `that is'. Hence it is crucial for the model to understand the difference between these terms to successfully generalize on the C2 test set. However, based on the train set, they both perform the same role: they act as a connector between two clauses in the command where the target needs to be identified based on the attributes in the first clause after satisfying the constraint of the following clause. We explain this more intuitively in Appendix \ref{app:unfair_expl}. To illustrate this empirically, we show that the average consistency between model predictions before and after replacing all `and' with `that is' in ReaSCAN test sets is 93.5.\footnote{Detailed results can be found in Table \ref{tab:c2_unfair}.} Hence, \emph{the train set of ReaSCAN is insufficient for the model to disambiguate between `and' and `that is', thereby rendering the C2 generalization task unfair.}

\textbf{Model Learns a Reasonable Alternate Hypothesis for C2.} We hypothesize that for the commands in the C2 split, the model treats the second `that is' as if it were `and', similar to the \texttt{2-relative-clause} commands it has seen in the train set. We consider this model behavior to be reasonable, since this hypothesis is consistent with the train set. To verify this empirically, we create a new set of examples, called C2-\texttt{alt}, by replacing the second `that is' with an `and' in all examples in the C2 test set. The model's predictions for C2 matched those for C2-\texttt{alt} 93\% of the time\footnote{Note that the model prediction for C2-\texttt{alt} matches the corresponding ground-truth for C2-\texttt{alt} 99\% of the time, showing that the model correctly handles those type of commands.}, clearly validating our hypothesis.

\textbf{Transformers Generalize to Higher Depths of Relative Clauses.} Since we showed that the C2 test set is unfair, we designed a new, fairer test split to check generalization capability of models to higher depths. We include commands up to depth 2 (i.e., the type of commands in the C2 test set) in the train set and test on commands of depth 3. We call this new test set as C2-\texttt{deeper}.\footnote{Details of this setup are provided in Appendix \ref{app:c2_correction}.} By including commands up to depth 2, we alleviate the issue of the model not being able to disambiguate between `that is' and `and'. Our best model achieved 85.6\% accuracy on C2-\texttt{deeper}. This is very surprising (since transformers have been known to struggle at depth recursion \cite{cogs}) and clearly shows that \emph{the multimodal transformer model is capable of generalizing systematically to higher depths} in this setting. This also re-affirms our claim that the original C2 test set was unfair.
\section{RefEx: Understanding How Transformers Ground and Compose}\label{sec:refex}

We wish to understand how transformers succeed on grounded compositional generalization tasks. However, the complexity of both, the ReaSCAN task, and the multi-modal transformer, makes it difficult to interpret the model. Hence, we design a new task, \textbf{RefEx}, based on the target identification subtask of ReaSCAN.\footnote{We discuss how RefEx differs from similar synthetic benchmarks in Appendix \ref{app:refex_diff}.} In the following sections, we show that even a one-layer, one-head attention-only model can successfully ground and compose multiple object attributes in RefEx. We then give a precise construction of the model, which demonstrates the detailed computations corresponding to grounding and composition.

\begin{figure}[t]
	\centering
	\includegraphics[scale=0.32]{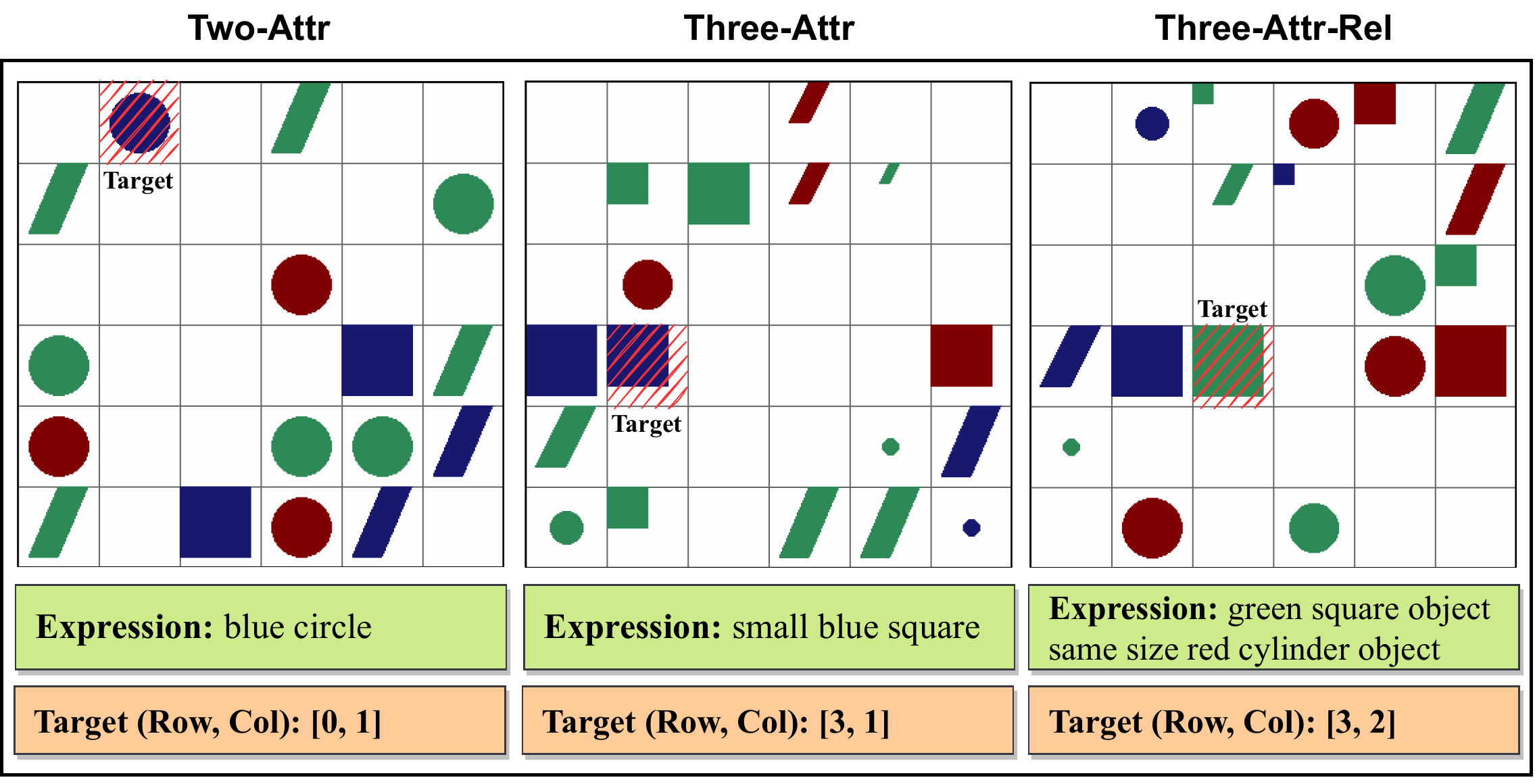}
	\caption{Examples from our RefEx dataset.}
	\label{ref_exp_task}
\end{figure}

\subsection{Task Setup and Test Splits}

Given a command that refers to a unique target object in the accompanying grid world, a model needs to identify the target location. Compared to ReaSCAN, etc., we get rid of subtasks like path planning and action sequence generation, and focus only on target identification.

We design three variants of the RefEx task with different command structures varying in difficulty:

\noindent(i) \texttt{two-attr}:=\texttt{\$\text{COL}} \texttt{\$\text{SHP}}. Model needs to ground and compose \emph{color} and \emph{shape} attributes. \\
\noindent(ii) \texttt{three-attr}:=\texttt{\$\text{SIZ}} \texttt{\$\text{COL}} \texttt{\$\text{SHP}}. Model needs to additionally handle the \emph{size} attribute, which requires relative reasoning. \\
\noindent(iii) \texttt{three-attr-rel}:=\texttt{\$\text{OBJ}} \texttt{\$\text{REL}} \texttt{\$\text{OBJ}}. Model effectively needs to perform two \texttt{three-attr} subtasks sequentially based on the relation between the referent and target objects.

\noindent Here, \texttt{\$\text{COL}} $\in \{\mathsf{red, green, blue}\}$, \texttt{\$\text{SHP}} $\in \{\mathsf{square, circle, cylinder}\}$, \texttt{\$\text{SIZ}} $\in \{\mathsf{small, big}\}$, \texttt{\$\text{OBJ}}:= \texttt{\$\text{SIZ}?} \texttt{\$\text{COL}?} \texttt{\$\text{SHP}?}, and \texttt{\$\text{REL}} $\in \{\mathsf{same\:size, same\:color, same\:shape}\}$. Figure \ref{ref_exp_task} shows examples from all three variants. Additionally, we create four compositional generalization test splits as described in Table \ref{tab:refex_splits}.

\begin{table}[t]
	\small{\centering
		\begin{tabular}{P{3em}p{18em}}
			\toprule
			\textsc{Split} & \textsc{Held-out Examples} \\
			\midrule
			A1 & \emph{green squares} as targets \\
			A2 & \emph{red circle} as targets or distractors \\
			A3\footnotemark & \emph{green circles} of size 2, referred with ``small'' \\
			A4\textsuperscript{\ref{foot:attr_1}} & command is ``small blue cylinder'' \\
			\bottomrule
		\end{tabular}
		\caption{\label{tab:refex_splits} Held-out examples in RefEx's compositional splits. Detailed descriptions are provided in Appendix \ref{app:refex_split_desc}.}
	}
\end{table}

\footnotetext{\label{foot:attr_1}Only for the \texttt{three-attr} variant.}

\begin{table}[t]
	\footnotesize{\centering
		\begin{tabular}{p{5.5em}P{3em}P{1.2em}P{1.2em}P{1.2em}P{1.2em}P{1.2em}}
			\toprule 
			\footnotesize{\textsc{Variant}} &
			\footnotesize{\textsc{Layers}} &
			\footnotesize{\textsc{R}} &
			\footnotesize{\textsc{A1\footnotemark}} & 
			\footnotesize{\textsc{A2}} &
			\footnotesize{\textsc{A3}} &
			\footnotesize{\textsc{A4}}
			\\
			
			\midrule
			
			\texttt{two-attr} &
			1 &
			100 &
			100 &
			100 &
			- &
			- 
			\\

			\midrule
			
			\texttt{three-attr} &
			1 &
			100 &
			100 &
			100 &
			100 &
			100
			\\

			\midrule
			
			\multirow{2}{=}{\texttt{three-attr-rel}} &
			1 &
			78.8 &
			31.9 &
			33.5 &
			- &
			-
			\\
			
			&
			2 &
			99.7 &
			99.4 &
			98.8 &
			- &
			-
			\\
			
			\bottomrule
		\end{tabular}
		\caption{\label{tab:refexp_main} Performance of attention-only transformer (with single attention head in each layer) on the RefEx task. \footnotesize{\textsc{R}} stands for random.}
	}
\end{table}

\subsection{Model}\label{simple_model}

We consider attention-only transformers (with residual connections) with two layers or less. We use natural sparse embedding matrices (see Fig. \ref{fig:emb_mx}, \ref{fig:emb_mx_three}, \ref{fig:emb_mx_three_rel}) to represent the command and world state. The input sequence length $n = 2 + 36$ for \texttt{two-attr} where $2$ and $36$ correspond to the number of command and grid world tokens, respectively. The output representations corresponding to the grid world tokens are mapped to logits by taking element-wise sum followed by softmax for 36-way classification (each class represents a unique grid location). See Appendix \ref{app:refex_details} for more details.

\subsection{Results and Discussion}

\footnotetext{\label{foot:mod_data}Result on modified training distribution. See Section \ref{app:refex_a1}.}

The performance of the model described above on the RefEx task is shown in Table \ref{tab:refexp_main}.

\noindent\textbf{Efficacy of Self-attention Layers.} For \texttt{two-attr}, we found it surprising to see that a one-layer, one-head attention-only transformer can successfully ground and compose the attributes for correct target identification. Moreover, the model generalizes to novel compositions of the attributes as can be seen from its performance on the compositional splits.

In \texttt{three-attr}, which is more difficult than \texttt{two-attr}, surprisingly, a one-layer, one-head model can ground and compose three different attributes, including \emph{size}, which requires complex relative reasoning.
%-----
Finally, in \texttt{three-attr-rel}, we find that at least two layers are required to solve the task. This makes intuitive sense: each layer will solve one \texttt{three-attr} subtask to identify the referent or target object.
%-----

\noindent\textbf{From RefEx to ReaSCAN.} Inspired by these results, we evaluate attention-only transformers on ReaSCAN target identification. Examples in ReaSCAN can contain up to three referring expression subtasks. Therefore, based on our intuition above, the model's performance should saturate after 3 layers. We show these results in Fig. \ref{fig:attn_only}. Moreover, since the design of splits in RefEx is similar to that of ReaSCAN, we were able to derive useful insights from models trained on RefEx in order to improve the performance on ReaSCAN A2 split. See Appendix \ref{app:refex_a1} for details.

\subsubsection{Interpreting how Transformers Ground and Compose}\label{sec:interpret}

We completely describe how an attention-only transformer with one attention layer and one head solves \texttt{two-attr} (\texttt{three-attr} and \texttt{three-attr-rel} are similar but more complex). 
Let's first recall how the one-layer, one-head model works. We denote the input token embeddings by $\mathbf{x}_1, \ldots, \mathbf{x}_n$ and the final output embeddings by $\mathbf{r}_1, \ldots, \mathbf{r}_n$. 
Let $\mathbf{W}_Q, \mathbf{W}_K, \mathbf{W}_V$ be the parameter matrices for queries, keys, and values; we can then define the query, key and value vectors for $i \in [n]$ by $\mathbf{q}_i = \mathbf{W}_Q \mathbf{x}_i$, $\mathbf{k}_i = \mathbf{W}_K \mathbf{x}_i$, and $\mathbf{v}_i = \mathbf{W}_V \mathbf{x}_i$. For each $i \in [n]$ we compute the output of attention block as
$\tilde{\mathbf{r}}_i = \mathbf{W}_o \sum_{j=1}^{n}\alpha_{i,j} \mathbf{v}_j$, where the attention scores are given by $(\alpha_{i, 1}, \ldots , \alpha_{i, n}) = 
\mathrm{softmax}(\langle \mathbf{q}_i, \mathbf{k}_1 \rangle, \ldots , \langle \mathbf{q}_i, \mathbf{k}_n \rangle)$. The residual connection then gives $\mathbf{r}_i = \tilde{\mathbf{r}}_i + \mathbf{x}_i$. As described in Section \ref{simple_model}, in our model, the final grid cell containing the target is chosen by applying softmax to the logits corresponding to 36 grid world tokens, $L_{n-36}, \ldots, L_n$. Let $\mathbf{1}$ be the all-ones vector; for \texttt{two-attr}, we have $\langle \mathbf{1}, \mathbf{x}_i \rangle = 2$ when $i$ corresponds to grid world tokens (see embedding matrix in Figure \ref{fig:emb_mx}). Now, we show the computation of logit $L_i$ where $i$ corresponds to grid world tokens.
\begin{align*}
L_i &= \langle \mathbf{1}, \mathbf{r}_i \rangle = \langle \mathbf{1}, \mathbf{x}_i \rangle + \langle \mathbf{1}, \tilde{\mathbf{r}}_i \rangle \\
 &= 2 + \langle \mathbf{1},  \mathbf{W}_o \sum_{j=1}^{n}\alpha_{i,j} \mathbf{v}_j \rangle \\
 &= 2 + \sum_{j=1}^{n} \alpha_{i,j} \langle \mathbf{1},  \mathbf{W}_o  \mathbf{v}_j \rangle \\
 &= 2 + \sum_{j=1}^{n} \alpha_{i,j} \mathbf{s}_j
\end{align*}

\begin{figure}[t]%
	\centering
	\subfloat[\centering Learned Model]{{\includegraphics[scale = 0.17]{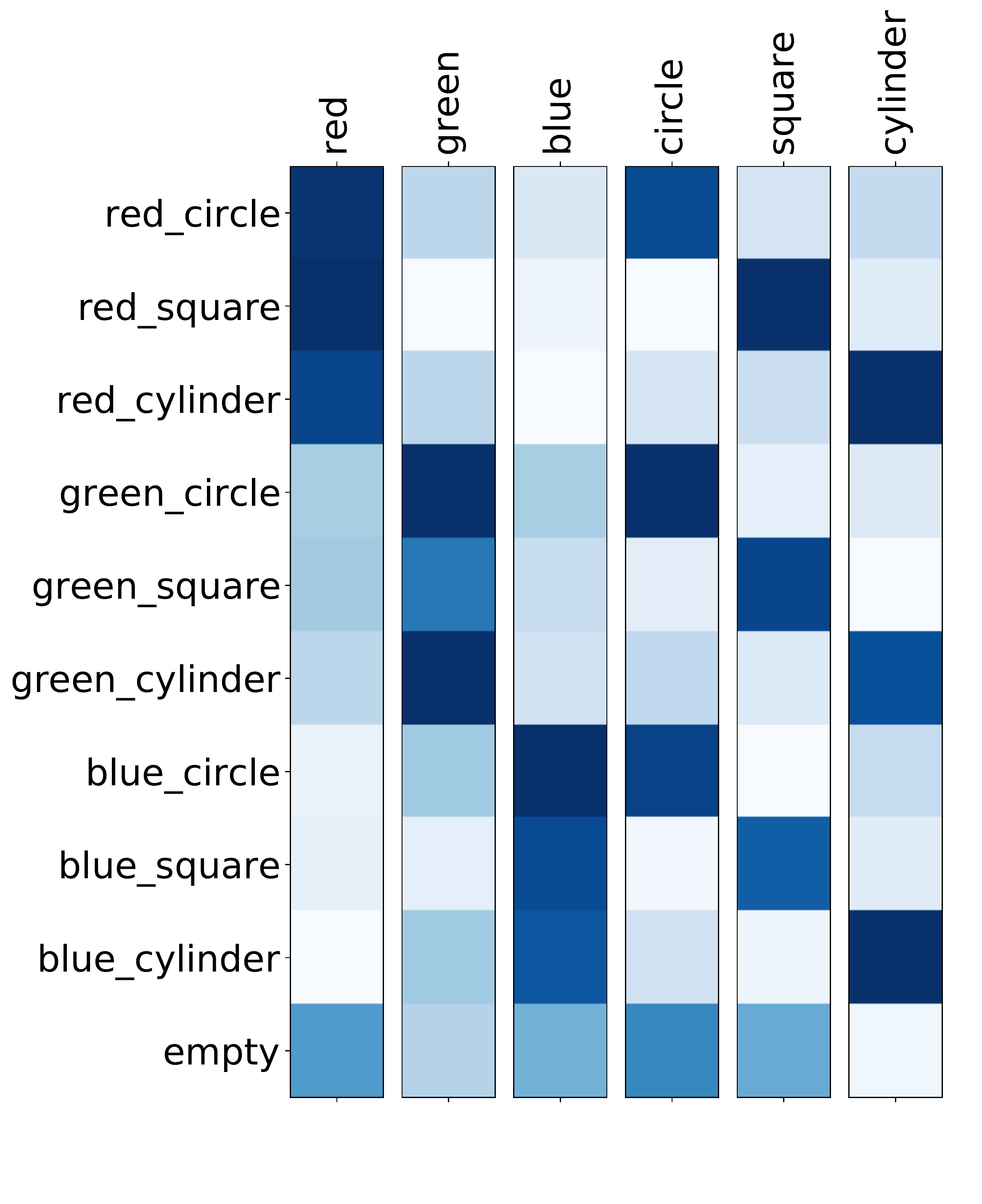} }}%
	%	\qquad
	\subfloat[\centering Our Construction]{{\includegraphics[scale = 0.17]{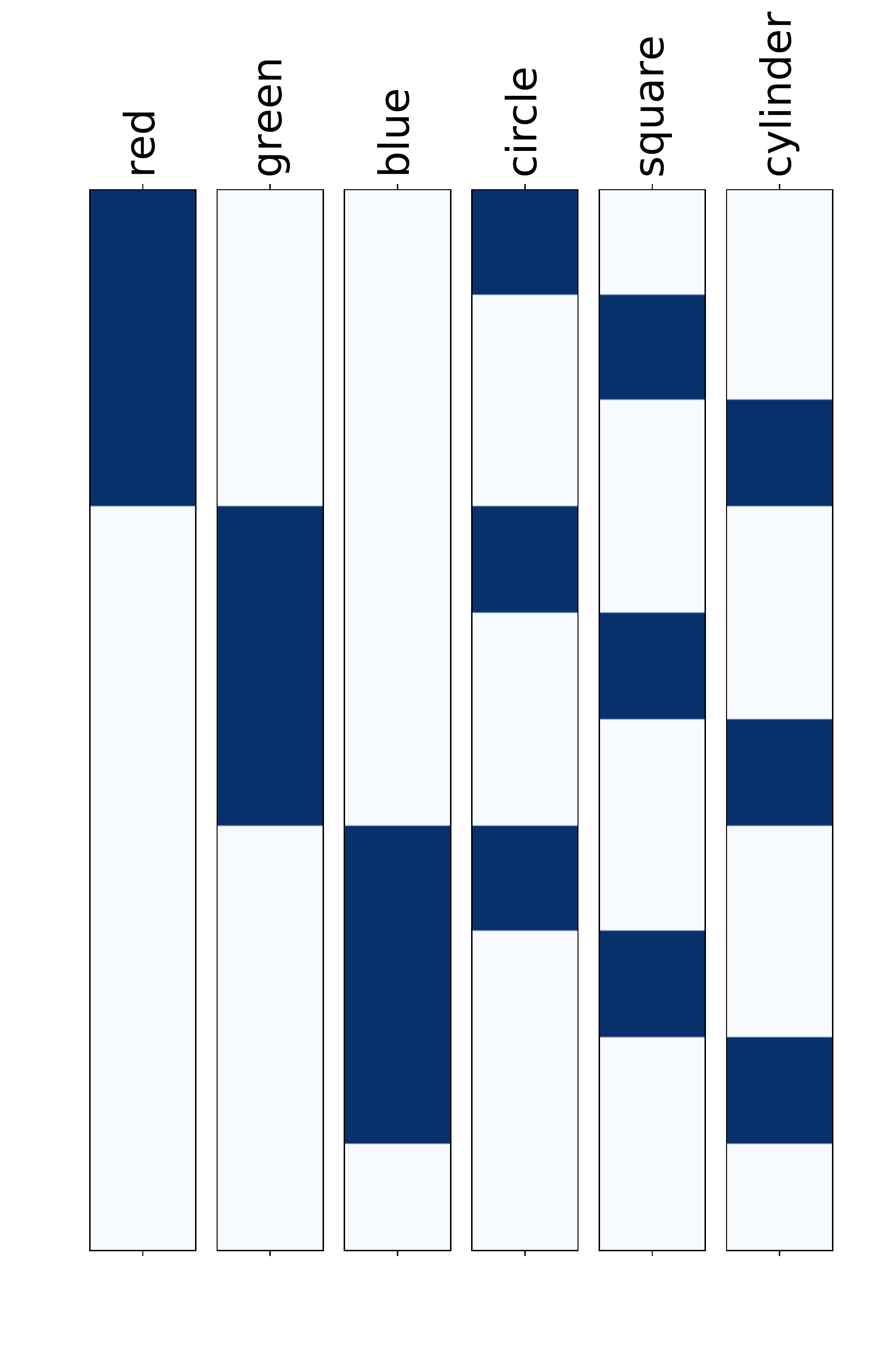} }}%
	
	\caption{(a) Portion of matrix $\mathbf{M}_\mathrm{Learned}$ for the attention-only transformer trained on the \texttt{two-attr} variant and (b) Portion from $\mathbf{M}_\mathrm{Construct}$ matrix (right side). Darker grid cells correspond to higher values in the matrix. Refer to Figure \ref{fig:qk_matrices_full} and \ref{fig:s_full} for full matrices.}%
	\label{fig:qk_matrices}%
\end{figure}

We now qualitatively illustrate on a specific example of \texttt{two-attr} and show how the learned parameters ($\mathbf{M}_\mathrm{Learned}$ in Figure \ref{fig:qk_matrices})) lead to the correct target prediction.
Note that the matrix $\mathbf{M} \in \mathbb{R}^{n_{\mathrm{vocab}} \times n_{\mathrm{vocab}}}$ here contains the dot product of query and key vectors of all possible pairs of tokens in the vocabulary. The rows in $\mathbf{M}$ correspond to queries while the columns correspond to keys.
Let the command tokens be $\langle \mathsf{red}\rangle, \langle \mathsf{cylinder}\rangle$.
We now show that the logit for $\langle \mathsf{red}\ \mathsf{cylinder}\rangle$ is the maximum among all possible values for grid tokens.
In the learned model, we observe that when $i$ and $j$ correspond to grid world tokens either $\alpha_{i,j}$ or $\mathbf{s}_j$ is very small :
\[\mathbf{L}_{i} \approx 2 + \sum_{j\in C} \alpha_{i,j} \mathbf{s}_j,\]
where $C$ is the set of indices of command tokens in our example, i.e., $C = \{1, 2\}$. 
When there is a match in an attribute in the grid world token $i$ (say $\langle \mathsf{red}\ \mathsf{circle} \rangle$) and a command token $j$ (say, $\langle \mathsf{red}\rangle$), then the corresponding summand in the above sum, i.e. $\alpha_{i,j} \mathbf{s}_j$ is large, and when there is no match (e.g., when the command token $j$ is changed to $\mathsf{green}$) then it is small. Therefore, in our example, the logit computation for the token $\langle \mathsf{red}\ \mathsf{cylinder}\rangle$ has two large summands and corresponds to a \emph{full match}, whereas for tokens like $\langle \mathsf{blue}\ \mathsf{cylinder} \rangle$ and $\langle \mathsf{red}\ \mathsf{circle} \rangle$, there is only one large summand. This corresponds to \emph{partial match}. Finally, for a token like $\langle \mathsf{green}\ \mathsf{square}\rangle$, there is \emph{no match} i.e. none of the summand, in this case, is large.
While the above argument is qualitative, by looking at the entries of $\mathbf{M}_{\mathrm{Learned}}$ it can be made quantitative; $\mathbf{M}_{\mathrm{Learned}}$ led us to construct
$\mathbf{M}_{\mathrm{Construct}}$ (Fig. \ref{fig:qk_matrices}(b)) which makes the computations transparent while staying faithful to the learned model.
Looking along the columns of the matrix corresponding to tokens $\langle \mathsf{red}\rangle$ and $\langle \mathsf{cylinder}\rangle$, we can see that the row corresponding to $\langle \mathsf{red}\ \mathsf{cylinder}\rangle$ grid world token has two ``dark'' cells (\emph{full match}), while rows corresponding to $\langle \mathsf{blue}\ \mathsf{cylinder} \rangle$, $\langle \mathsf{red}\ \mathsf{circle} \rangle$ and $\langle \mathsf{green}\ \mathsf{square}\rangle$ have at most one ``dark'' cell (i.e. \emph{partial match} or \emph{no match}). Thus the grid cell corresponding to $\langle \mathsf{red}\ \mathsf{cylinder}\rangle$ will be the model's output. \emph{Full match} here corresponds to the idea that both visual attributes $\langle \mathsf{red}\rangle$ and $\langle \mathsf{cylinder}\rangle$ mentioned in the command, were successfully grounded to the grid world token $\langle \mathsf{red}\ \mathsf{cylinder}\rangle$, and the composition of these two successful groundings contributed to two large summands in the final logit computation.

Finally, empirically our constructions attain perfect accuracies across all splits.
The matrices of our construction for \texttt{three-attr} are in Figure \ref{fig:three_attr_m_c} and \ref{fig:three_attr_s_c}.

\section{Related Works}

\textbf{Compositional Generalization.} Modern deep learning models perform extremely well on in-distribution test sets. However, unlike humans, they fail at generalizing compositionally \cite{scan, cogs, cfq}. Recent works have investigated the compositional generalization abilities of models in grounded setups using datasets such as CLEVR \cite{clevr}, CLOSURE \cite{closure}, and gSCAN \cite{gscan}. In this work, we focus on the task setup of gSCAN, and additionally work with GSRR \cite{google_multimodal} and ReaSCAN \cite{reascan}. Both these works propose new challenging splits for gSCAN. Prior works have proposed many different specialized methods for solving gSCAN \cite{gao, kuo, heinze}. Similar to \citet{devil, revisiting}, in this work, we show that even simple transformer-based models, with minor modifications to the architecture or training data distribution, perform well on the task and serve as strong baselines for future work.

\textbf{Probing and Interpreting Models.} In this work, we used a linear probe \cite{belinkov-probing} to analyze the target identification abilities of models. Similar to our work, \citet{rasp, transformercircuits, attention_norm} and many others also attempt to explain the inner workings of transformers, possibly for non-synthetic problems. Unlike most of these works, however, our construction essentially \emph{completely} describes the computations of the learned models. To the best of our knowledge, we are the first to study \emph{how} self-attention facilitates compositional generalization in grounded environments.

\section{Conclusion and Future Work}

Recent benchmarks like gSCAN and ReaSCAN test grounded compositional generalization abilities of ML models. In this work, we identify key modifications in multimodal transformers that improve compositional generalization on these benchmarks. 
%------
With a battery of probing experiments, we found that identifying the target location is the main challenge for the models. Additionally, we showed that a particular test set in ReaSCAN is \emph{unfair} and proposed a modified train-test split in its stead. 
Finally, we designed a new task, \textbf{RefEx}, to study grounding and composition in attention-only transformers. We showed the efficacy of \emph{single} self-attention layer with \emph{single} head in successfully grounding and composing multiple visual attributes in a grid world environment. From the learned models, we derived an explicit and interpretable construction that captures the model's behavior and completely describes the detailed computations corresponding to grounding and composition.

While our focus has been on \emph{tabula rasa} models, it is also of interest to see if pretraining on large datasets enables good performance on the considered benchmarks. Our preliminary 
investigations on GPT-3 and Codex (Appendix \ref{app:llms}) suggest that these text-based models have some way to go; more thorough investigation is left for the future.

We expect future work to address the current limitations of models on action sequence side compositional generalization, i.e., generalizing to novel combinations of action tokens. Moreover, our results indicate that designing compositional generalization splits can be surprisingly subtle and require careful scrutiny. Finally, grounded compositional generalization benchmarks should also target more realistic setups with natural images in the future.

\section{Limitations}

Our proposed approach fails on some gSCAN splits, specifically D, G, and H. These splits are designed to test output sequence-side systematic generalization capabilities of the model. In the future, we intend to extend our model by making architectural changes on the decoder side in order to tackle these splits. 

On ReaSCAN, our approach achieves 86.1\% on the B2 split and 76.3\% on the C1 split, which is relatively less compared to the performance on other splits, suggesting room for further improvement. We expect that the C1 split also has similar issues like the C2 split (see section \ref{sec:issue_c2}), although we haven't yet succeeded at concretely identifying these issues.

In line with previous models, our model also uses grid cell encodings which are very simple and explicitly represent different attributes of the world objects. However, natural images are high-dimensional and contain entangled representations of object attributes. Ideally, we would like to evaluate the compositional generalization abilities of models in a real-world setting using natural images since that has direct applications.

Our constructions for attention-only transformers in Section \ref{sec:refex} are given only for the RefEx task which is a relatively simpler task than ReaSCAN. We plan to give similar constructions for more complex tasks than RefEx in the future.

\section*{Acknowledgements}

We thank the anonymous reviewers for their constructive comments and suggestions. We would also like to thank Satwik Bhattamishra for reviewing an early draft of the paper and providing valuable feedback. We are grateful to our colleagues at Microsoft Research India for engaging in useful discussions and constantly supporting us throughout this work.

\bibliography{citations}
\bibliographystyle{acl_natbib}

\clearpage
\newpage
\appendix

\section{Model Architecture}\label{app:model_architecture}

\begin{figure*}[t]
	\centering
	\includegraphics[scale=1.2]{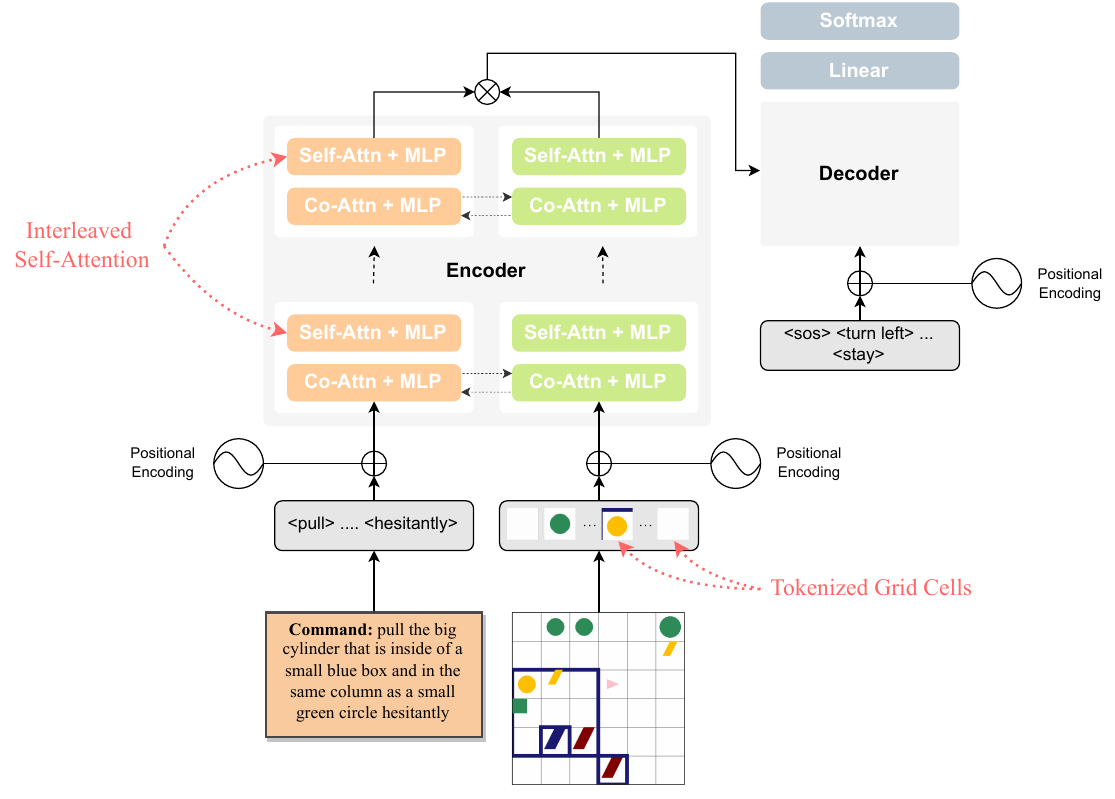}
	\caption{\label{fig:model} Illustration of the architecture for our proposed GroCoT model.}
\end{figure*}

The model architecture of our proposed approach GroCoT is shown in Figure \ref{fig:model}.

\section{Implementation Details}\label{app:implement}

\begin{table*}[t]
	\scriptsize{\centering
		\begin{tabular}{p{17.5em}P{11em}P{11em}P{11em}}
			\toprule
			\textbf{Hyperparameters} &\textbf{ReaSCAN} &\textbf{GSRR} &\textbf{gSCAN} \\ 
%			\cmidrule(lr){2-3}\cmidrule(lr){4-5}\cmidrule(lr){6-6}
%			\textbf{Hyperparameters} & Transformer & LSTM & Transformer & LSTM & Transformer \\
			\midrule
			\# Self-attention Layers (Vision) & 6 & 3 & 3 \\
			
			\# Self-attention Layers (Language) & 6 & 3 & 3 \\
			
			\# Co-attention Layers & 6 & 3 & 3 \\

			\# Decoder Layers & 6 & 6 & 6 \\
			
			Embedding Size & 128 & 128 & 128 \\
			
			Hidden/FFN Size & 256 & 256 & 256 \\
			
			Attention Heads & 8 & 8 & 8 \\
			
			Learning Rate & [0.00005, 0.00008, \textbf{0.0001}] & [0.00005, 0.00008, \textbf{0.0001}] & [\textbf{0.00005}, 0.00008, 0.0001] \\
			
			Batch Size & [\textbf{32}, 64] & [32, \textbf{64}] & [32, \textbf{64}] \\
			
			Dropout & 0.1 & 0.1 & 0.1 \\
			
			\midrule
			
			%			\# Parameters & 8.5M & 130M & 15M & 140M & 16M & 143M \\
			\# Parameters & 4.5M & 3M & 3M  \\
			
			Epochs & 100 & 100 & 100 \\
			
			Avg Time (Overall) & 64 & 14 & 18 \\
			
			\bottomrule
		\end{tabular}
		\caption{\label{tab:hyperparams} Different hyperparameters considered for models trained on ReaSCAN, GSRR, and gSCAN. Best hyperparameters for each model are in bold. (Avg Time shown above is in hours).}
	}
\end{table*}

We use PyTorch \cite{pytorch} for all our implementations. All our models were trained from scratch, and the parameters were updated using Adam optimizer. We designed a compositional validation set by taking few examples from each compositional splits of the respective dataset. The best model is selected based on the accuracy on this compositional validation set. Hyperparameter tuning was done using grid search. We show the best hyperparameters for our models corresponding to different datasets in Table \ref{tab:hyperparams}. Moreover, we show the average performance of 3 different runs with random seeds for all the models in the paper. We used 8 NVIDIA Tesla V100 GPUs each with 32 GB memory for all our experiments.

\section{Details of Datasets}\label{app:datasets}

\subsection{ReaSCAN}\label{app:reascan}
ReaSCAN consists of around 500K train, 30K validation, and 6K test examples where each example is a pair of command and world state. Given these two as input, models are supposed to output the correct sequence of action tokens. Apart from the above splits, ReaSCAN also has 7 systematic generalization test splits in total.
ReaSCAN has three types of input commands.

\begin{itemize}
	\item \texttt{Simple}:= \texttt{\$\text{VV}} \texttt{\$\text{ADV}?} (equivalent to gSCAN commands) 
	\item \texttt{1-relative-clause}:= \texttt{\$\text{VV}} \texttt{\$\text{OBJ}} \text{that}\ \text{is}\ \texttt{\$REL\_CLAUSE} \texttt{\$\text{ADV}?}
	\item \texttt{2-relative-clauses}:= \texttt{\$\text{VV}} \texttt{\$\text{OBJ}} \text{that}\ \text{is}\ \texttt{\$REL\_CLAUSE} \text{and}\  \texttt{\$REL\_CLAUSE} \texttt{\$\text{ADV}?}
\end{itemize}

\begin{figure*}[!t]
	\begin{minipage}{\textwidth}
		
		\begin{minipage}{0.7\textwidth}
			\centering
			\resizebox{\linewidth}{!}{%
				\centering
				\begin{tabular}{rll}
					\toprule
					\texttt{Syntax} & \textbf{Descriptions} & \textbf{Expressions} \\ \midrule
					\texttt{\$VV} & verb & \{walk to, push, pull\} \\
					\texttt{\$ADV} & adverb & \{while zigzagging, while spinning, \\
					& & cautiously, hesitantly\} \\
					\texttt{\$SIZE} & attribute & \{small, big\}$^{*}$ \\
					\texttt{\$COLOR} & attribute & \{red, green, blue, yellow\} \\
					\texttt{\$SHAPE} & attribute & \{circle, square, cylinder, box, object\} \\ [1ex]
					\texttt{\$OBJ} & objects & (a | the) \texttt{\$SIZE?} \texttt{\$COLOR?} \texttt{\$SHAPE} \\
					\texttt{\$REL} & relations &  \{in the same row as, in the same column as, \\ && in the same color as,
					in the same shape as, \\ && in the same size as, inside of\} \\ 
					\texttt{\$REL\_CLAUSE} & clause & \texttt{\$REL} \texttt{\$OBJ} \\
					\bottomrule
			\end{tabular}}
			\captionof{table}{Definitions of syntax used in ReaSCAN command generation.${}^{*}$the actual size of any shape is chosen from \{1,2,3,4\} as in gSCAN \cite{gscan}. This table is taken from \citet{reascan}.}\label{tab:ReaSCAN-DSLs}
		\end{minipage}
		\hfill
		\begin{minipage}{0.28\textwidth}
			\centering
			\captionsetup[subfloat]{labelformat=empty}
			\begin{tabular}{cc}
				\small same row as &  \small same column as \\
				\includegraphics[width = 0.60in]{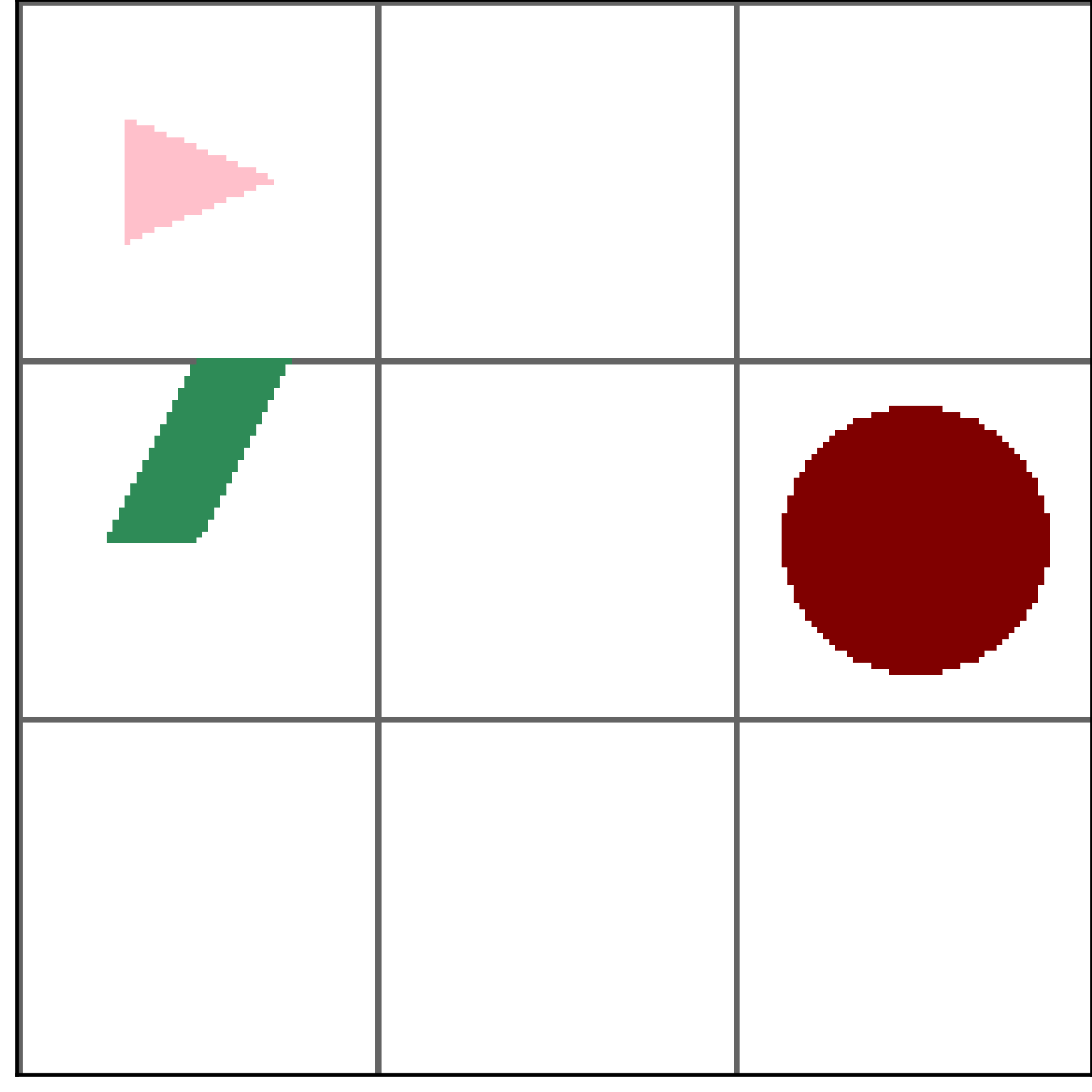}  &
				\includegraphics[width = 0.60in]{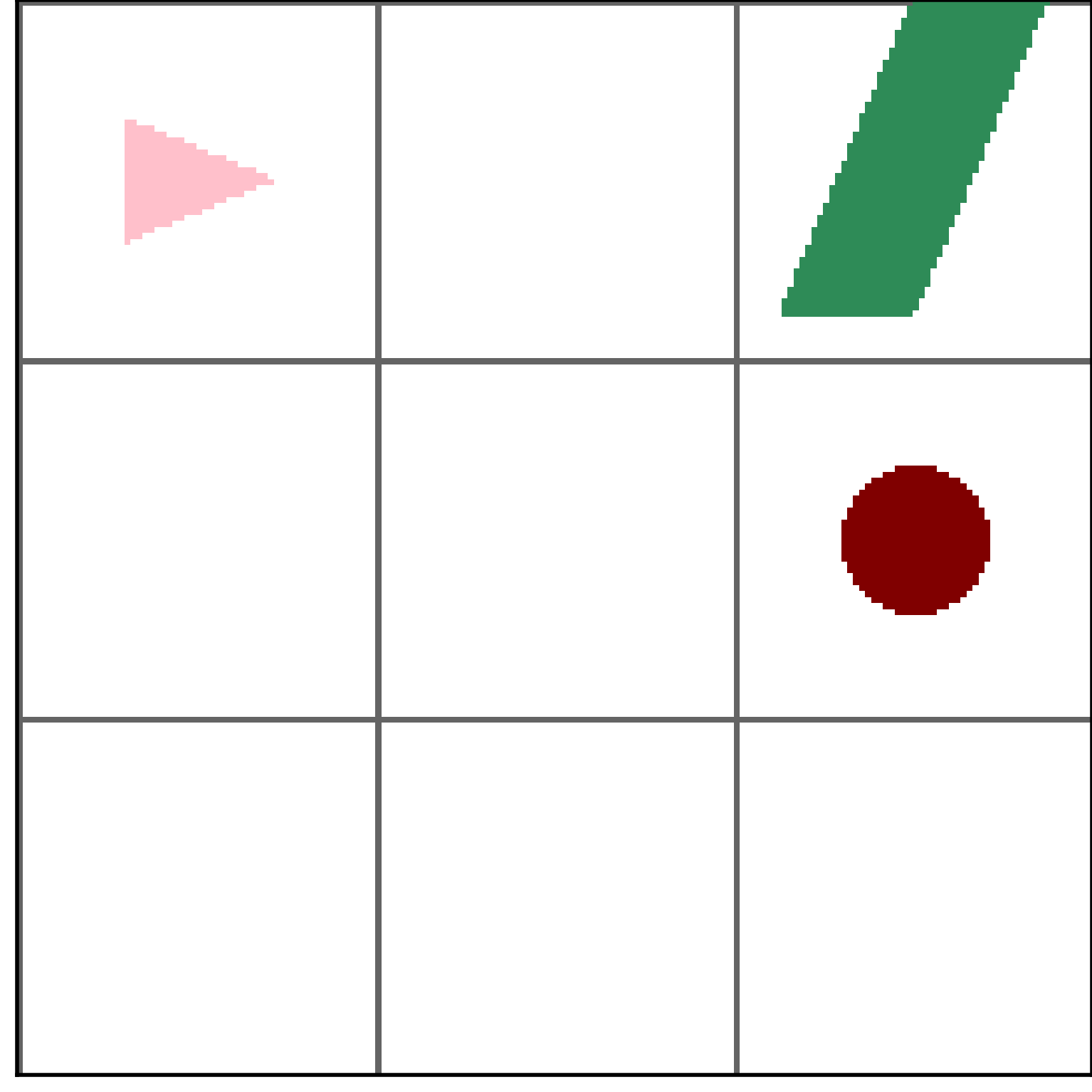} \\
				\small same color as &  \small same shape as \\
				\includegraphics[width = 0.60in]{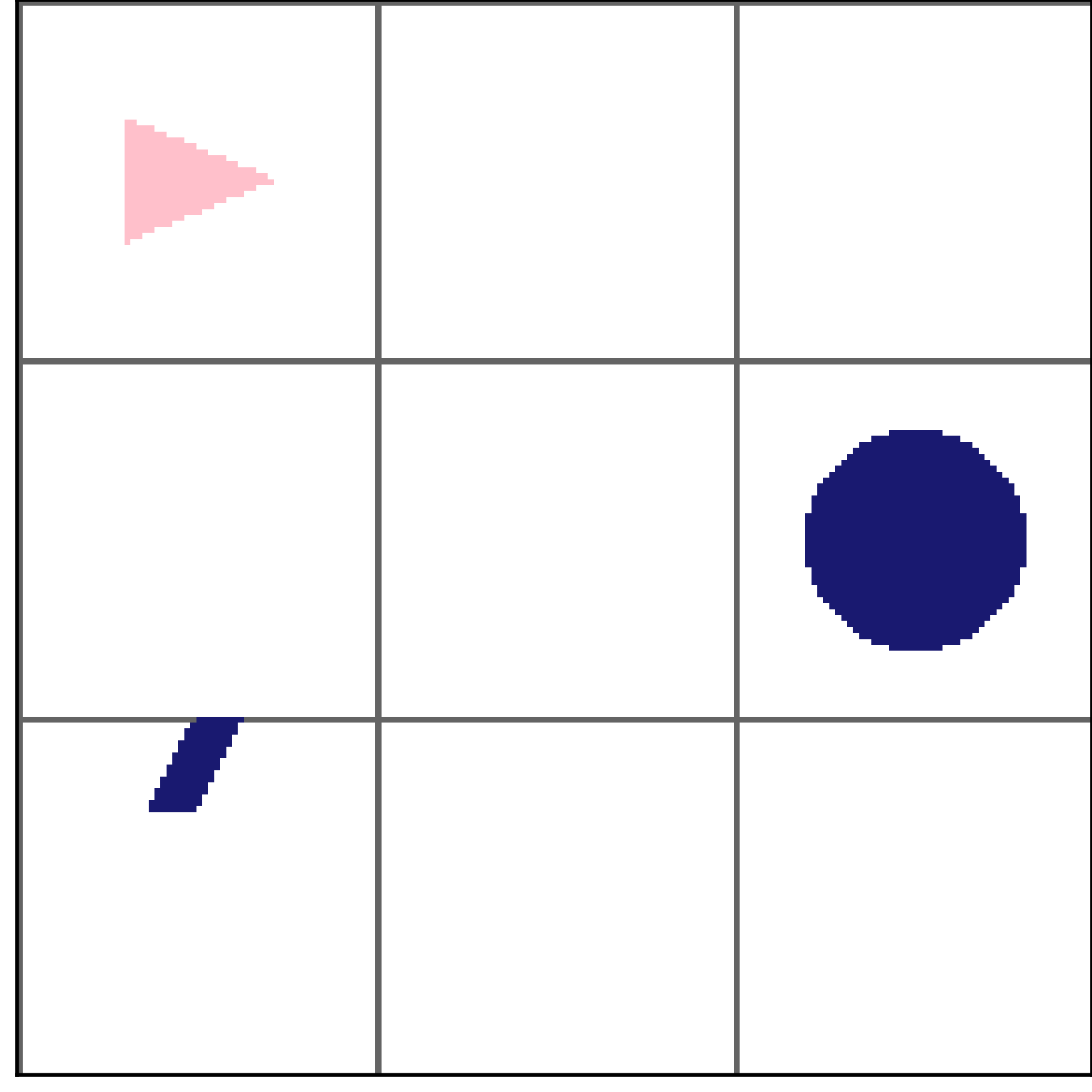} &
				\includegraphics[width = 0.60in]{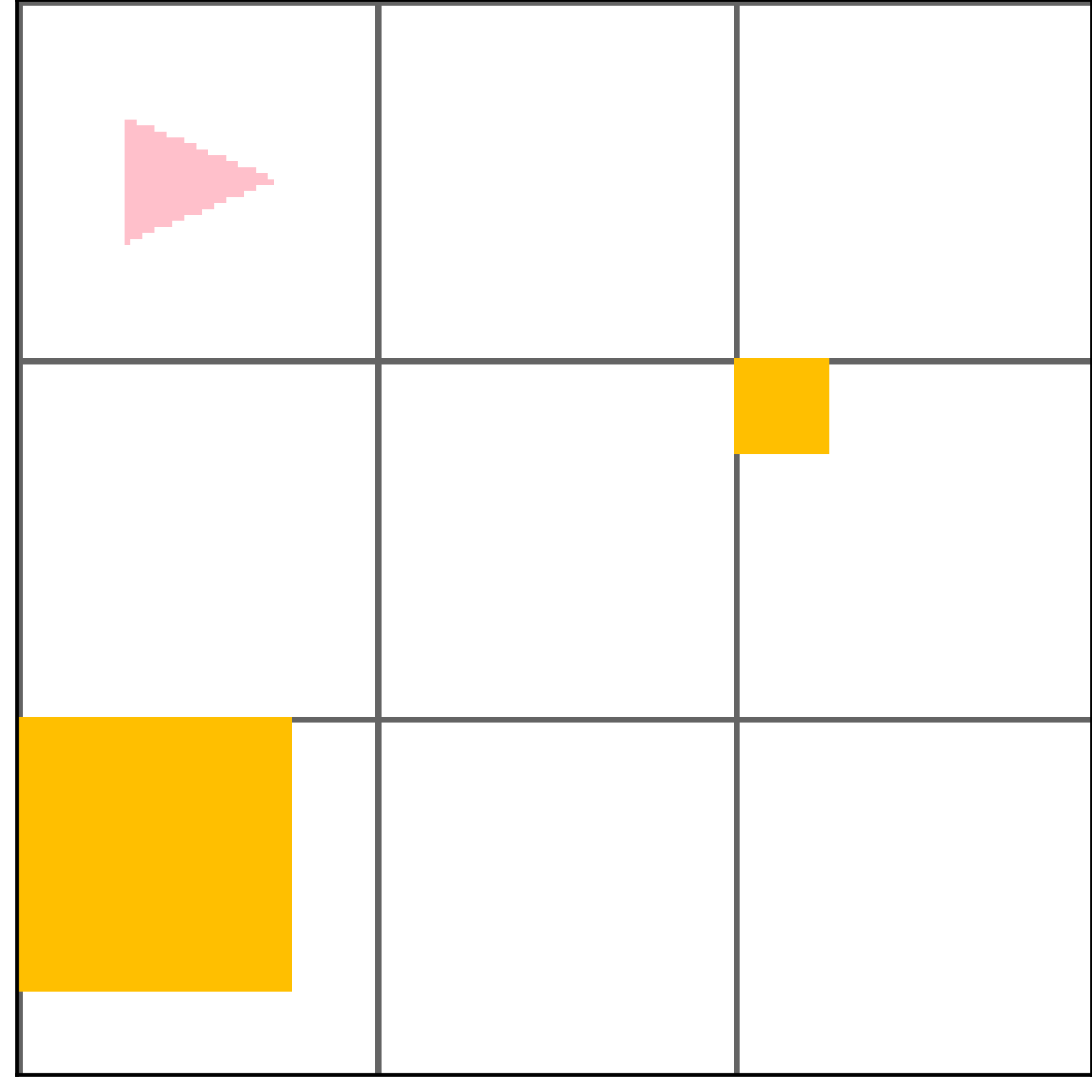} \\
				\small same size as &  \small inside of \\
				\includegraphics[width = 0.60in]{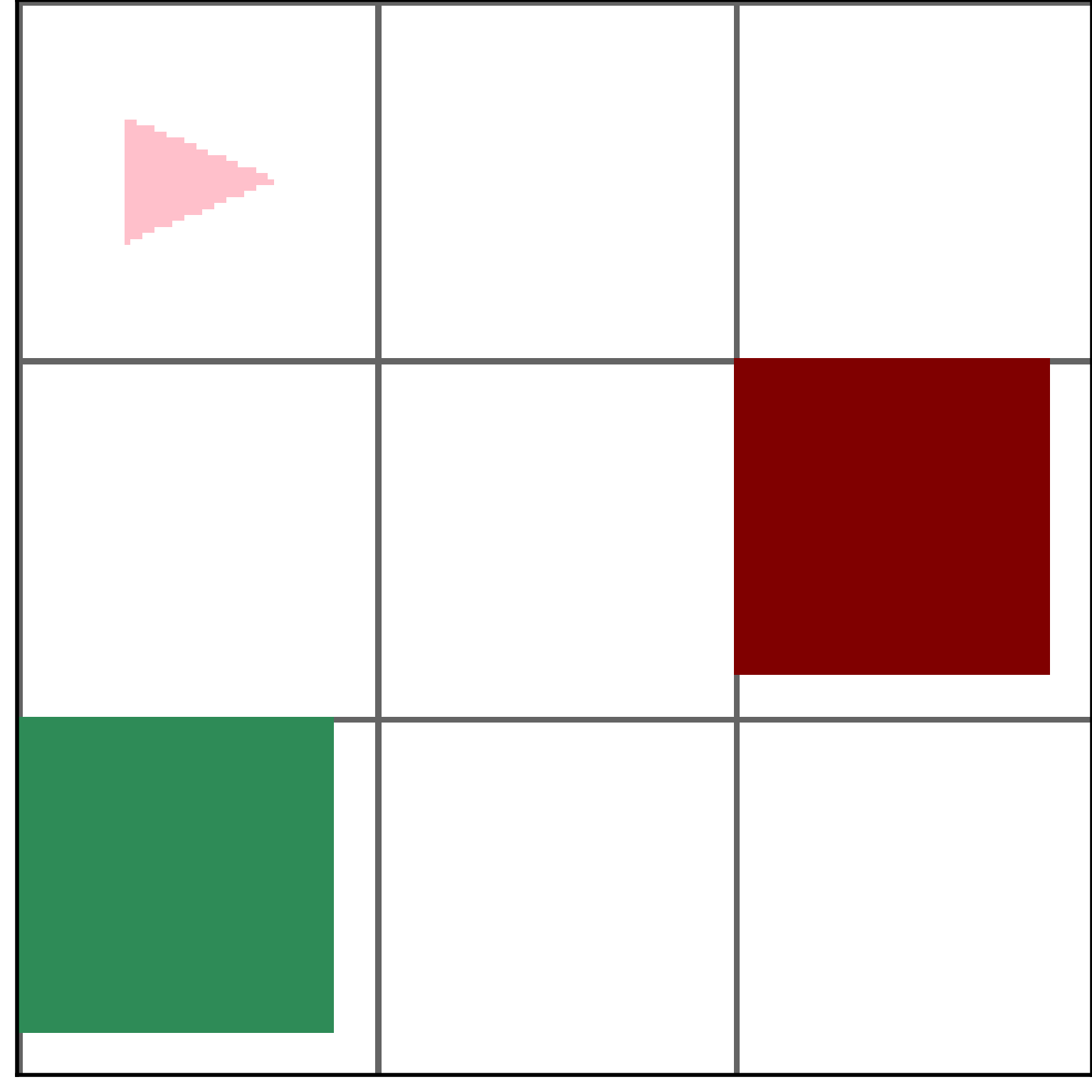} &
				\includegraphics[width = 0.60in]{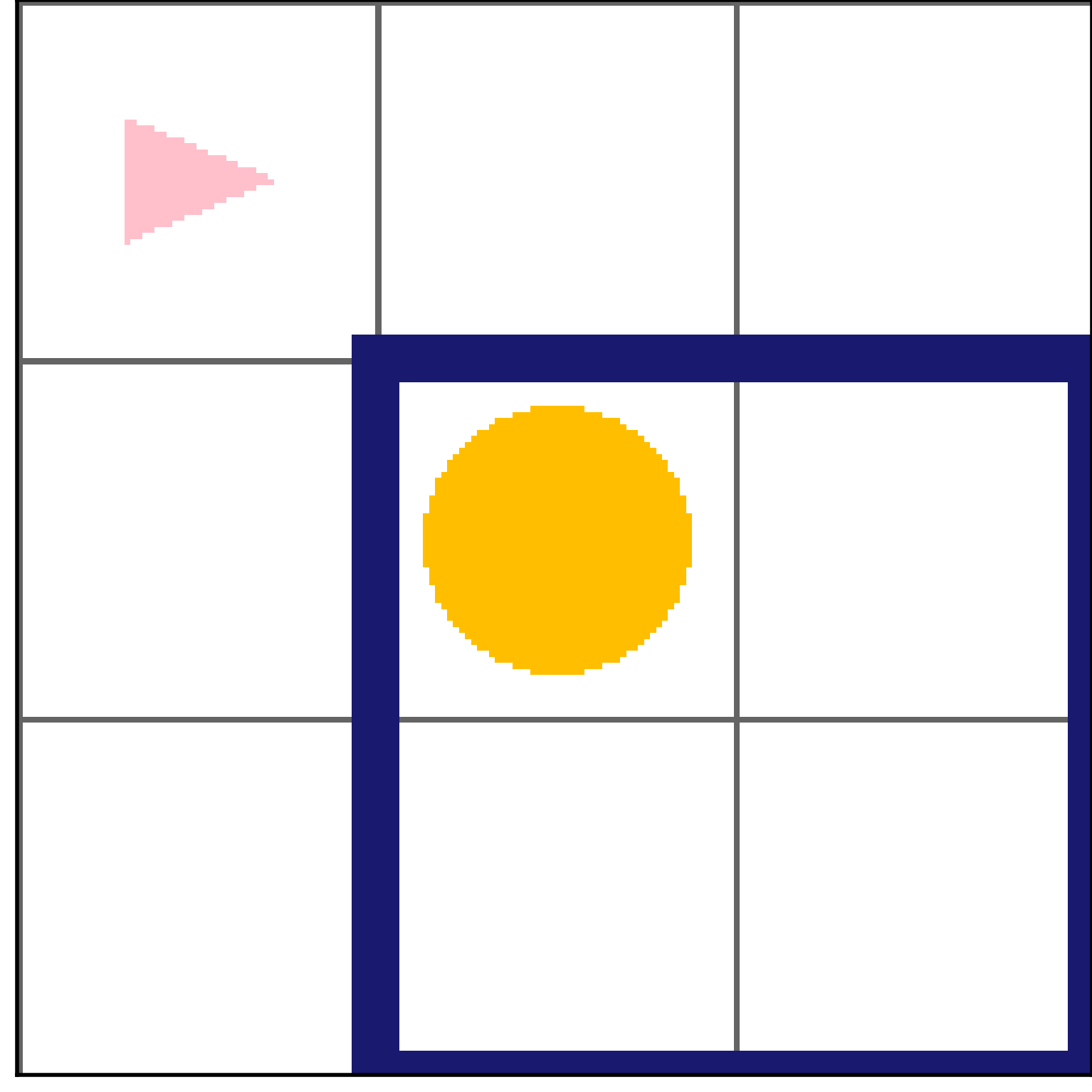} \\
			\end{tabular}
			\captionof{figure}{Relations.}
		\end{minipage}
		
	\end{minipage}
\end{figure*}

See Table \ref{tab:ReaSCAN-DSLs} for expansions of non-terminals in the grammar. Below, we describe the ReaSCAN test splits in detail:

\noindent \textbf{A1} \emph{Novel Color Modifier}: For this split, the train set never contains ``yellow square'' in the command. However, commands containing expressions such as ``yellow circle'' and ``blue square'' are present in the training set. During test time, this split expects model to zero-shot generalize on the phrase ``yellow square''.

\noindent \textbf{A2} \emph{Novel Color Attribute}: Here, the examples in which red squares are targets are held out from the train set. Commands in the train set also never contain the phrase ``red square''. However, the train set may contain red square objects as distractors in the background. This particular split tests model performance on novel combination of target object's visual attributes.

\noindent \textbf{A3} \emph{Novel Size Modifier}: In this split, a particular combination of size and shape attributes is held out from the train set. Specifically, the model never sees phrases like ``small cylinder'' or ``small green cylinder'' during training. While testing, the model needs to generalize to examples where cylinders of any color are referred using the ``small'' size attribute. Additionally, size being a relative concept in ReaSCAN, adds another level of complexity. For instance, size ``small'' can refer to an object of size 2 in a particular example, and in another example size ``small'' can instead refer to an object of size 3, depending on the other objects present in the grid world.

\noindent \textbf{B1} \emph{Novel Co-occurrence of Objects}: In this split, the commands contain those objects which never co-occur in training (e.g. ``small red circle'' and ``big blue square''). However, models do encounter these objects co-occurring with other objects during training. In summary, this split tests whether models can generalize over novel co-occurrence of objects.

\noindent \textbf{B2} \emph{Novel Co-occurrence of Relations}: Commands containing both ``same size as'' and ``inside of'' relations are held out from the training data. At test time, models must generalize to the novel co-occurrence of these two relations. Importantly, during training model does encounter commands where the relation ``inside of'' co-occurs with other relations except ``same size as''. 

\noindent \textbf{C1} \emph{Novel Conjunctive Clause Length}:  This split contains commands with additional conjunctive clause i.e. the commands contains 3 relative clauses (e.g. ``push the small red circle that is in the same column as a big green square and inside of a big blue box and in the same row as a blue square hesitantly''). Models trained with up to 2-relative clauses must generalize to longer commands which contain 3 relative clauses.

\noindent \textbf{C2} \emph{Novel Relative Clauses}: In ReaSCAN train data, commands contain a maximum of 1 recursive relative clause (e.g. ``push the circle that is in the same column as a yellow square and inside of a big box cautiously''). However, this test split contains commands with 2-recursive relative clauses i.e. there are two ``that is'' clauses in the commands (e.g. ``push the blue circle that is in the same column as a blue cylinder that is in the same row as a green square hesitantly'').

\subsection{gSCAN}\label{app:gscan}
gSCAN is very similar to ReaSCAN; both are essentially grounded navigation tasks. The grid world in gSCAN is the same as that of ReaSCAN, although commands in ReaSCAN are much more complicated than gSCAN. Commands in ReaSCAN contain relative clauses, whereas gSCAN commands have no relative clauses. For example command from gSCAN looks like \emph{walk to a red big square}. gSCAN contains around 350K training and 20K test examples for the compositional splits. Below, we briefly describe the individual compositional test splits in gSCAN:

\noindent \textbf{A} \emph{Random}: This test split contains random examples and is supposed to test in-distribution generalization.

\noindent \textbf{B} \emph{Yellow squares}: For this split, the train set doesn't contain examples where \emph{yellow squares} are referred with color and shape attributes.

\noindent \textbf{C} \emph{Red Squares}: Here, the examples where the target is \emph{red square} are held out from the train set.

\noindent \textbf{D} \emph{Novel Direction}: For this split, the examples where the target objects are located south-west of the agent, are held out from training set.

\noindent \textbf{E} \emph{Relativity}: To create this split, examples where circles with size 2 are referred as \emph{small} are held out from the train set.

\noindent \textbf{F} \emph{Class Inference}: In this split, all examples where the verb is \emph{push}, and target is a square of size 3 are held out from the training set. Note that the model needs to infer that this object is of class `heavy' based on the size 3 and needs to push twice to move an object by one grid cell.

\noindent \textbf{G} \emph{Adverb k=1}: Only one example with the adverb \emph{cautiously} is shown in the train set.

\noindent \textbf{H} \emph{Adverb to Verb}: For this split, examples where the commands have verb \emph{pull} and adverb \emph{while spinning} are held out from the train set.

\begin{table*}[t]
	\small{\centering
		\begin{tabular}{p{15.1em}P{1.5em}P{8em}P{1.5em}P{1.5em}P{1.5em}P{1.5em}P{1.5em}P{1.5em}P{1.5em}}
			\toprule
			\small{\textsc{Model}} & \small{\textsc{A}} & \small{\textsc{Comp. Average}} & 
			\small{\textsc{B}} &
			\small{\textsc{C}} &
			\small{\textsc{D}} &
			\small{\textsc{E}} &
			\small{\textsc{F}} &
			\small{\textsc{G}} &
			\small{\textsc{H}}
			\\
			
			\midrule
			
			Multimodal LSTM 
			\scriptsize{\cite{reascan}} &
			97.7 &
			32.7 &
			54.9 &
			23.5 & 
			0.0 & 
			35.0 &
			92.5 &
			0.0 &
			22.7
			\\
			
			GCN-LSTM
			\scriptsize{\cite{gao}} &
			98.6 &
			%			62.3\todo{?} &
			- &
			99.1 &
			80.3 & 
			\textbf{0.2} & 
			87.3 &
			99.3 &
			- &
			\textbf{33.6}
			\\
			
			Multimodal Transformer \scriptsize{\cite{google_multimodal}} &
			\textbf{99.9} &
			60.0 &
			99.9 &
			99.3 & 
			0.0 & 
			99.0 &
			99.9 &
			0.0 &
			22.2
			\\

			GroCoT (ours) &
			\textbf{99.9} &
			\textbf{60.4} &
			\textbf{99.9} &
			\textbf{99.9} & 
			0.0 & 
			\textbf{99.8} &
			\textbf{99.9} &
			0.0 &
			22.9
			\\
			
			\bottomrule
		\end{tabular}
		\caption{\label{tab:gscan_result} Performance of GroCoT on gSCAN \cite{gscan}.}
	}
\end{table*}

\subsection{GSRR}\label{app:gsrr}
Based on the gSCAN task setup, \cite{google_multimodal} proposed 5 additional compositional generalization splits that also test spatial reasoning capabilities of models. GSRR contains around 250K train examples, from which specific kind of examples are held-out to create the systematic splits. Below, we describe the compositional splits in GSRR:

\noindent \textbf{\uppercase\expandafter{\romannumeral1\relax}} \emph{Random}: This split contains random examples for testing in-distribution generalization.

\noindent \textbf{\uppercase\expandafter{\romannumeral2\relax}} \emph{Visual}: For this split, the train set doesn't contain examples where \emph{red squares} are present either as targets or referent object.

\noindent \textbf{\uppercase\expandafter{\romannumeral3\relax}} \emph{Relation}: Here, the examples which contain both \emph{green squares} and \emph{blue circles} are held-out from the training data.

\noindent \textbf{\uppercase\expandafter{\romannumeral4\relax}} \emph{Referent}: For this split, the examples where \emph{yellow squares} are referred as target are held-out from the trainset.

\noindent \textbf{\uppercase\expandafter{\romannumeral5\relax}} \emph{Relative Position 1}: Here, all examples where the targets are north of their referent objects are held-out from the training set.

\noindent \textbf{\uppercase\expandafter{\romannumeral6\relax}} \emph{Relative Position 2}: In this split, all examples where the target is south-west of the referent object are not seen in training data.
\section{Performance of Large Language Models}\label{app:llms}

\begin{figure}[t]
	\centering
	\includegraphics[scale=0.3]{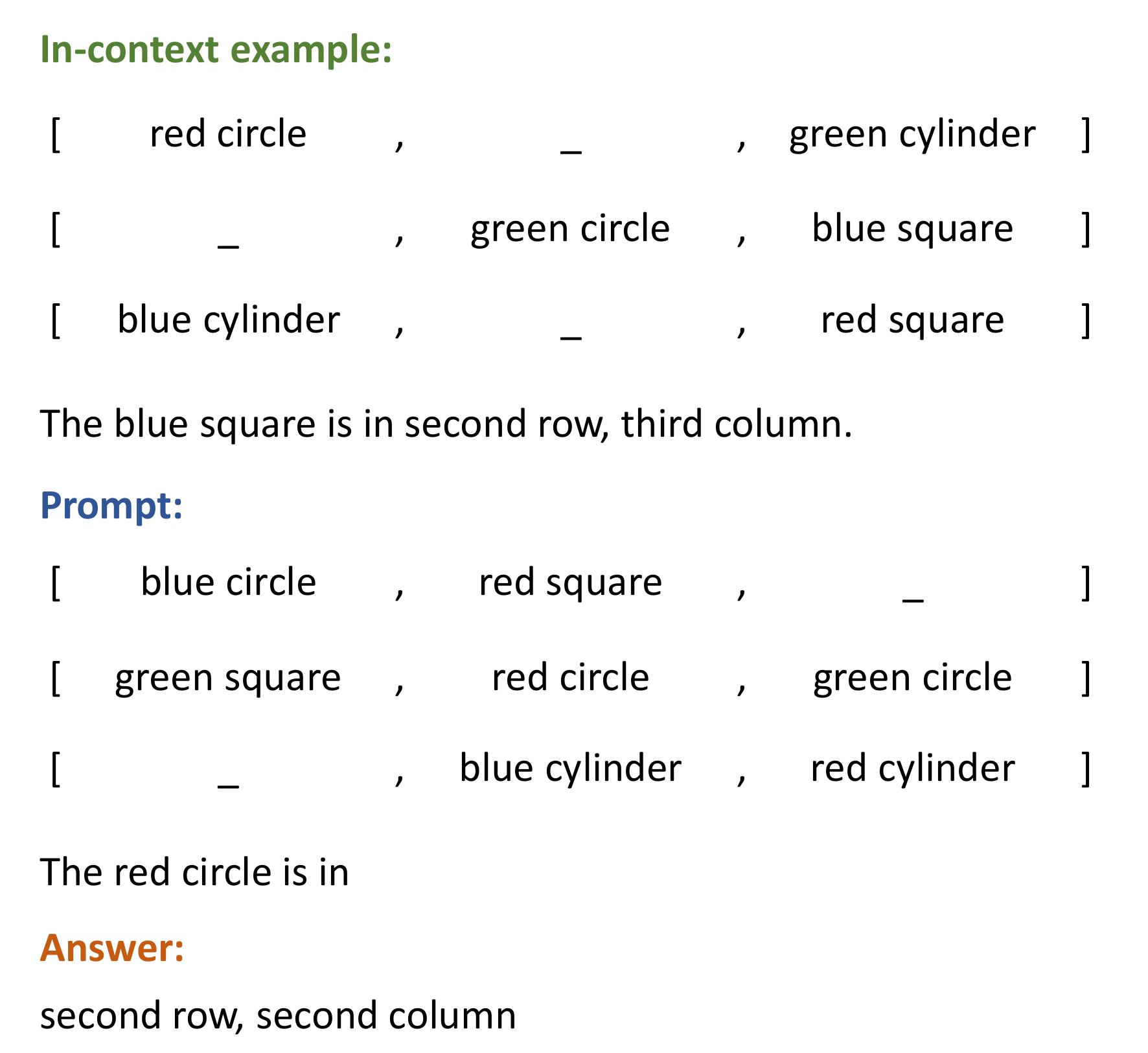}
	\caption{\label{fig:llm_direct} An example from the \texttt{direct-grounded} experiment on LLMs.}
\end{figure}

\begin{figure}[t]
	\centering
	\includegraphics[scale=0.3]{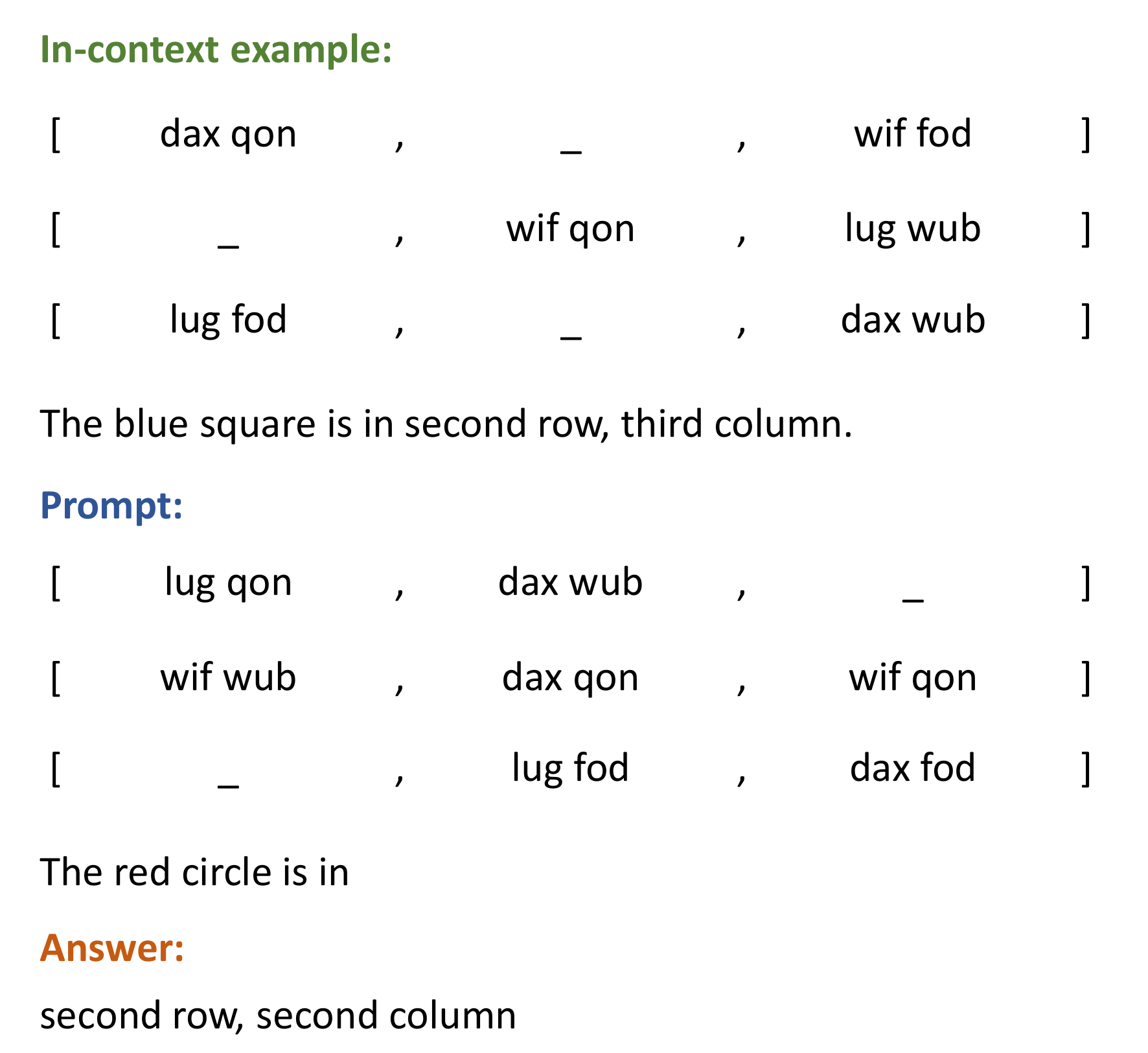}
	\caption{\label{fig:llm_nonsense} An example from the \texttt{nonsense-grounded} experiment on LLMs.}
\end{figure}

\begin{figure}[t]
	\centering
	\includegraphics[scale=0.3]{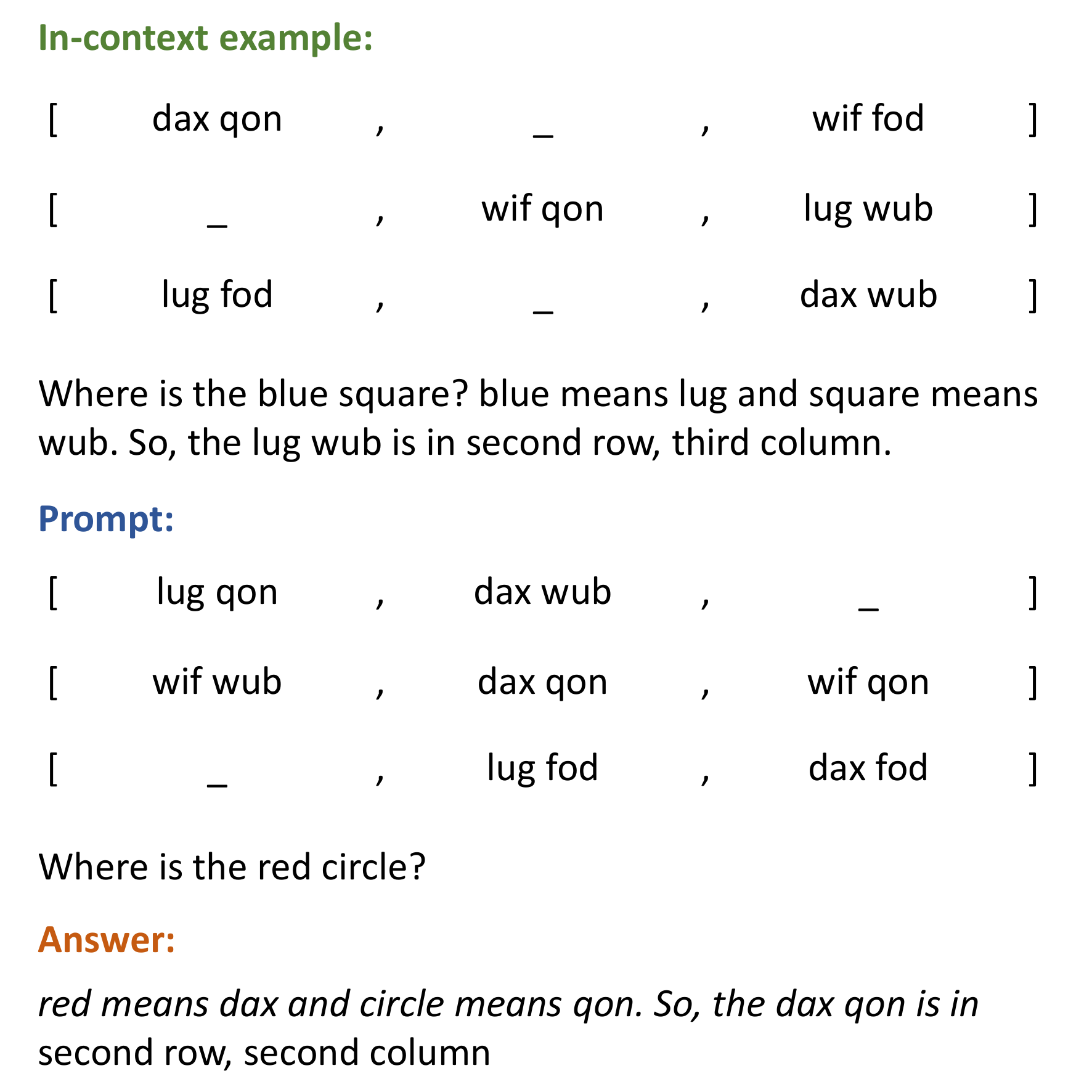}
	\caption{\label{fig:llm_cot} An example from the \texttt{chain-of-thought} experiment on LLMs.}
\end{figure}

We design a simpler version of the task to evaluate the performance of large language models (LLMs) such as GPT-3 \cite{gpt3} and Codex \cite{codex}. Given a $3\times3$ grid world and a simple command stating an object's color and shape, the model needs to output that object's location. The world state as well as the output are in a textual description format to make the task compatible with the input and output space of LLMs. Along with each test example, the models are given prompts containing multiple in-context examples. Note that the context provided to the model is ensured to contain all necessary information that the model might need to answer the test example.

In the most basic version of the experiment, called \texttt{direct-grounded}, we directly refer to the objects in the grid world with their attributes used in the command. See Fig. \ref{fig:llm_direct} for an illustration. We provide 30 in-context train examples to the model as part of the prompt for each of the 20 test example. In this setup, Codex achieved 95\% accuracy while GPT-3 achieved 65\% accuracy. This experiment merely serves as a sanity-checking baseline for the other experiments we described next.

Ideally, the model should learn the mappings of words in the commands to the tokens in the world state. Hence, in this experiment, called \texttt{nonsense-grounded}, we use non-sense words to refer to the objects in grid world as shown in Fig. \ref{fig:llm_nonsense}. This task setting more closely resembles the target identification task in ReaSCAN (while being a much more simpler version of it). In this setup, the models fail badly. Codex achieves only 25\% accuracy, and GPT-3 achieves only 30\%. This clearly shows that LLMs are as yet unable to tackle such grounded compositionality tests, even when provided with sufficient evidence via in-context training examples in the prompt.

Following recent work \cite{cot}, we provide explicit \emph{chain-of-thoughts} to the LLMs to make them understand the task. While from a purely evaluation point-of-view, this can be considered \emph{cheating}, we were merely curious to check whether the chain-of-thought idea, which has led to so much success in reasoning tasks, would help the models do better in this task setting. An illustration of the chain-of-thought provided to the models is given in Fig. \ref{fig:llm_cot}. Codex performs much better when provided with such chain-of-thoughts, achieving 70\% accuracy. However, GPT-3 still struggles on the task, achieving only 25\% accuracy.
\section{Additional Details and Results of Analysis Experiments}\label{app:appendix_analysis}

\subsection{Target Identification vs Navigation: What is the Challenge?}\label{app:target_loc}

\subsubsection{Comparison with Other Methods}\label{app:target_loc_prev_work}

Studies with similar objectives to ours have been carried out by \citet{gscan} and \citet{google_multimodal}. However, the results of their experiments are not very conclusive. \citet{gscan} examined the object in the grid world that was being maximally attended by the agent and checked whether it is the target. However, their results do not correlate well with the conclusions. For instance, they find that in the error cases on the gSCAN `C' split, the agent attends over the correct target object half the time! \citet{google_multimodal} compare the final position of the agent with the ground-truth target location. However, such an analysis fails to disentangle the subtasks of target identification and agent navigation and discards the causal relationship between them. Also, because of the interpretation of verbs such as \emph{push} and \emph{pull}, the final position of the agent may be very different from the target location.

\subsubsection{Predicting Target Location from Earlier Encoder Representations}

\begin{figure}[t]
	\centering
	\includegraphics[scale=0.54]{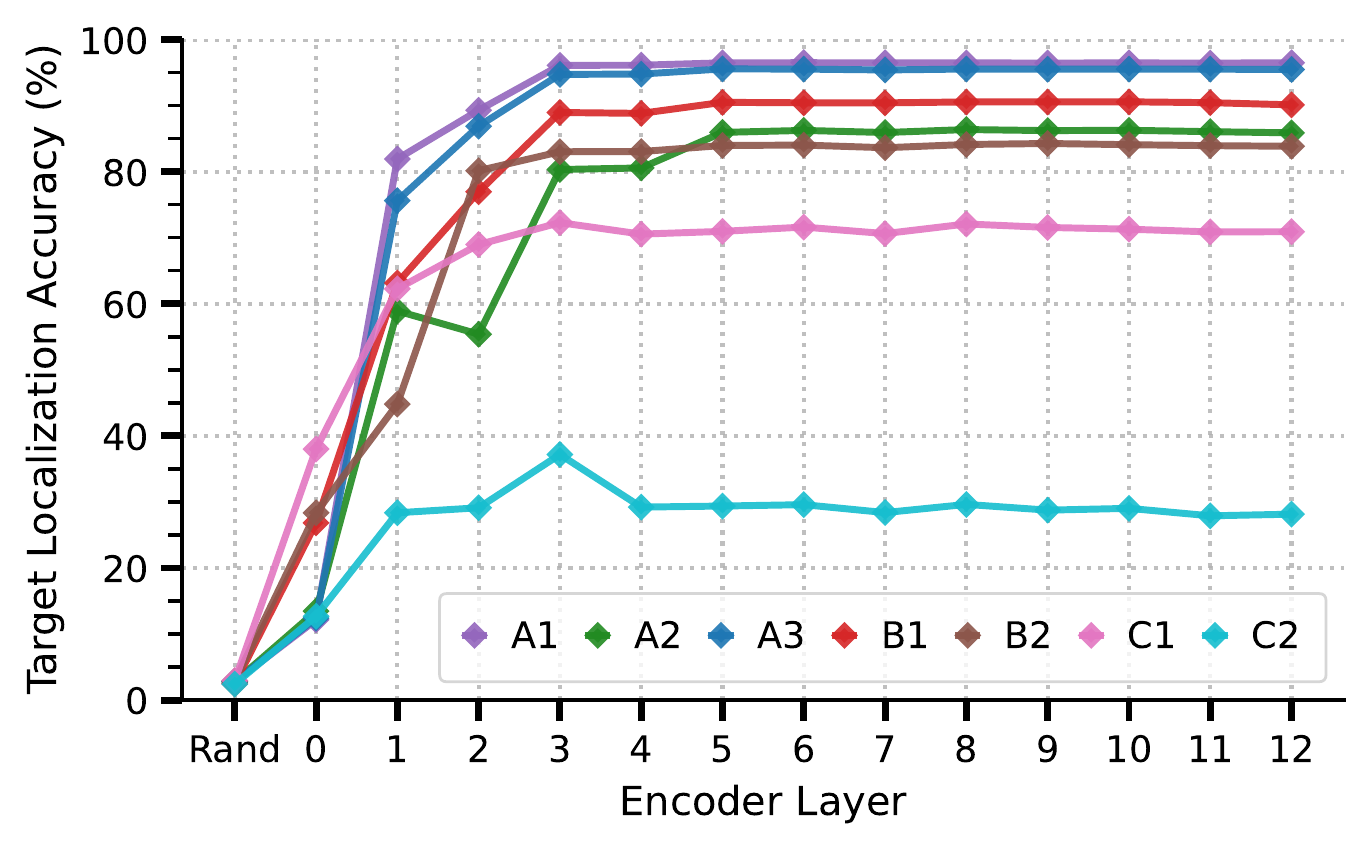}
	\caption{\label{fig:linear_probe} Probing target location information in encoder representations at different layers. Layer 0 corresponds to directly training a linear layer over the input embeddings and Rand corresponds to training a linear layer on random embeddings.}
\end{figure}

We experimented with predicting the target location from earlier encoder representations, including the embeddings and randomly initialized representations. The latter two act as baselines for our probing results described in section \ref{sec:target_loc}. The results are provided in Figure \ref{fig:linear_probe}.

\subsubsection{Target Identification Results on gSCAN}\label{app:target_loc_gscan}

We show that target identification is the main challenge for most of the gSCAN splits as well. Similar to the experiment described in section \ref{sec:target_loc}, we experiment with providing ground-truth target locations to the model. As seen from the results provided in Table \ref{tab:target_loc_gscan}, identifying the target location is the main challenge for most of the splits in gSCAN.

\begin{table}[t]
	\footnotesize{\centering
		\begin{tabular}{P{5em}P{5em}}
			\toprule
			\small{\textsc{Split}} & \small{\textsc{Accuracy}} \\
			
			\midrule
			
			B & 99.9 \\
			C & 100 \\
			D & 0.0 \\
			E & 100 \\
			F & 99.9 \\
			G & 0.0 \\
			H & 23.4 \\
			
			\bottomrule
		\end{tabular}
		\caption{\label{tab:target_loc_gscan} Results of GroCoT on gSCAN when provided with ground-truth target locations in the input.}
	}
\end{table}

\subsection{Issues in ReaSCAN Test Set Design}\label{app:unfair}

\subsubsection{Intuitive Explanation of Unfairness in C2 Split Design}\label{app:unfair_expl}

In this section, we explain why the model would not be able to disambiguate between \emph{`and'} and \emph{`that is'} based on the train set. Since the meanings of \emph{`and'} and \emph{`that is'} are apparent to humans, to understand things from the model's perspective, let us replace them with non-sense words \emph{`axyo'} and \emph{`tafyo'} respectively. The model sees commands such as ``walk to the small red square \textbf{tafyo} in the same row as the big blue cylinder'' and ``walk to the small red square \textbf{tafyo} in the same row as the big blue cylinder \textbf{axyo} in the same column as a green circle'' during training. There isn't enough information here to make the model understand that \emph{`tafyo'} applies the constraint on its right to the clause on its immediate left while \emph{`axyo'} applies the constraint on its right to the clause that occurs on the immediate left of the \emph{`tafyo'} before it. Looking at these examples, both \emph{`tafyo'} and \emph{`axyo'} do the exact same thing, i.e., apply the constraint on their immediate right over the first clause in the command.

\begin{table}[t]
	\footnotesize{\centering
		\begin{tabular}{P{3em}P{5em}P{5em}P{5em}}
			\toprule
			Splits & Action Seq Accuracy (Default) & Action Seq Accuracy (``and'' replaced) & Consistency
			\\
			
			\midrule
			
			A1 &
			99.40 &
			99.26 &
			99.61
			\\
			
			A2 &
			88.74 &
			88.41 &
			96.86
			\\
			
			A3 &
			98.13 &
			97.54 &
			98.90
			\\
			
			B1 &
			93.55 &
			94.43 &
			97.36
			\\
			
			B2 &
			86.15 &
			87.19 &
			95.33
			\\
			
			C1 &
			74.06 &
			68.90 &
			72.78
			\\
			
			\bottomrule
		\end{tabular}
		\caption{\label{tab:c2_unfair} Measuring the consistency of the  predictions made by the model before and after replacing all \emph{and}'s with \emph{that is}'s in ReaSCAN.}
	}
\end{table}

\subsubsection{Experimental Details of Evaluation on C2-deeper}\label{app:c2_correction}

We randomly select 100,000 examples from the train set and 6,000 examples from the C2 test set. This forms the new train set. We generate 4500 new examples of depth three to form the C2-\texttt{deeper} test set.
\section{Additional Details and Results on our RefEx Task}\label{app:refex}

\subsection{Differences with Other Benchmarks}\label{app:refex_diff}

RefEx dataset is different from existing similar-looking synthetic benchmarks like SHAPES \cite{nmn}, CLEVR \cite{clevr}, and CLEVR-Ref+ \cite{clevref}. RefEx aims to test the systematic generalization capabilities of neural models in a grounded setting while keeping the task simple enough to allow easier interpretation of the model's behaviour. On the contrary, previous diagnostic benchmarks are more concerned with testing the overall reasoning capabilities of the model. Also, the close resemblance between RefEx and ReaSCAN allows us to use insights gained from the RefEx task to improve performance on ReaSCAN.

\subsection{Details about the Task and Model}\label{app:refex_details}

\subsubsection{Details About the Task}\label{app:refex_split_desc}

Each variant in RefEx contains 90K training, 2.5K validation, and 2.5K test examples.

\noindent \textbf{A1} We hold out all examples where the command contains ``green square''. As a result, the model never sees a \emph{green square} object as the target although green squares occur in the background in the train set. This split expects the model to zero-shot generalize over the composition of ``green'' and ``square'' attributes.

\noindent \textbf{A2} We hold out all examples where the command contains ``red circle'' while ensuring that model never encounters \emph{red circle} object during training.

\noindent \textbf{A3}\footnote{\label{foot:attr_2}Only for the \texttt{three-attr} variants.} We hold out all examples where the command is ``small green circle'' and the corresponding target is of size 2, meaning that the model has never seen a green circle of size 2 being referred to with ``small''. 

\noindent \textbf{A4}\textsuperscript{\ref{foot:attr_2}} We hold out all examples where the command is ``small blue cylinder''. At test time, the model needs to zero-shot compose the concept of ``small'' with ``blue cylinder'' objects.

\subsubsection{Details About the Model}\label{app:refex_model}

\begin{figure}[t]
	\centering
	\includegraphics[scale=1.2]{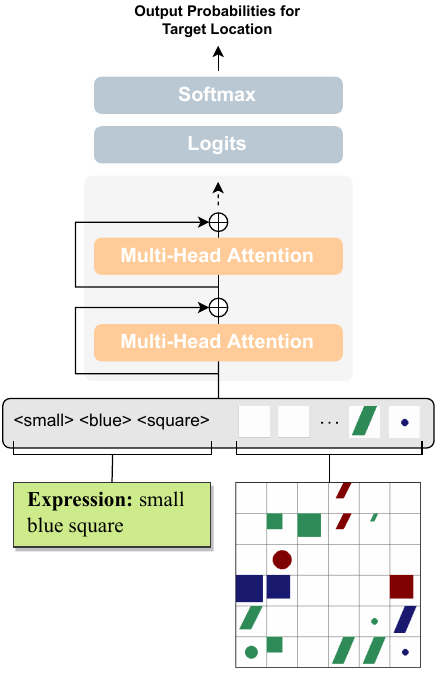}
	\caption{\label{fig:refex_model} Model architecture used for the RefEx experiments.}
\end{figure}

 Below, we describe the details of our attention-only transformer. We begin by mapping the command and the world state to $d_{\mathrm{model}}$ size embeddings using our embedding matrix. These embeddings constitute the initial input $\mathbf{X} \in \mathbb{R}^{d_{\mathrm{model}} \times n}$ to our network, where $n$ is the sequence length. Here, $n = 2 + 36$ for \texttt{two-attr} variant, $n = 3 + 36$ for \texttt{three-attr} variant, and $n = 8 + 36$ for \texttt{three-attr-rel} variant. Here, $2$, $3$, and $8$ correspond to the tokens in the referring expression. Note that this input representation $\mathbf{X}$ contains information from both modalities. This representation is fed into the first multi-head attention block, and the subsequent outputs are residually added back to the initial input. 
% forming $\mathbf{x_1} = \mathbf{X} + \sum_{h \in H}h(\mathbf{x_0})$
We repeat this mechanism for successive attention blocks, and after the $n^{th}$ attention block, the final representations corresponding to the world state tokens are mapped to logits by taking element-wise sum along the $d_{\mathrm{model}}$ dimension. In the end, we apply softmax operation on the logits for 36-way classification, where each class corresponds to a particular grid cell. The architecture is illustrated in Figure \ref{fig:refex_model}.
%i.e. $\mathbf{x_n^W} \in \mathbb{R}^{d_{model} \times 36}$ 

The only learnable parameters for our model come from the query, key, value, and output matrices of different attention layers. For \texttt{two-attr} and \texttt{three-attr} variants, we don't require positional information for the command while we incorporate learned positional embeddings for the \texttt{three-attr-rel} variant.

\begin{figure}[htbp]
	\centering
	\includegraphics[scale=0.5]{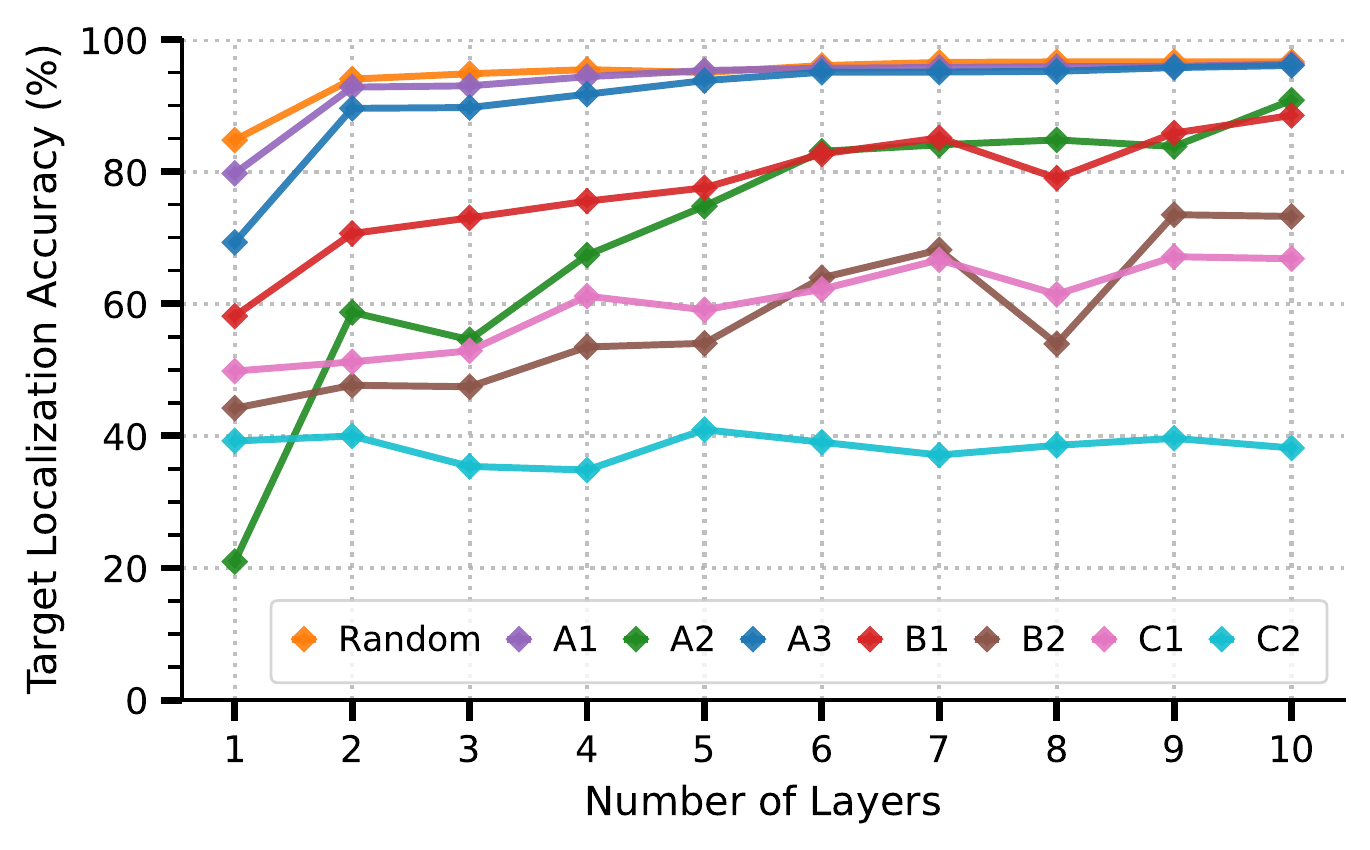}
	\caption{\label{fig:attn_only} Target Localization accuracy for attention-only transformer on ReaSCAN. The model's performance plateaus after 3 layers for most of the compositional splits.}
\end{figure}

\subsection{Understanding Model Performance on the RefEx A1 Split}\label{app:refex_a1}

When we first created the Referring Expressions train and test sets, the attention-only transformer achieved perfect generalization on all splits except A1. On A1, the model achieved average accuracy. This was very surprising because the model was able to solve A2, which seems a strictly harder task than A1. The only difference between the two splits is that in A1, the held-out target object (\emph{green square}) is seen as a distractor while in A2, the object (\emph{red circle}) is never seen as a distractor. One probable hypothesis we had for the model failing on A1 was that the model was overfitting on the fact that \emph{green squares} are always distractors in the train set, thereby preventing the model to predict it as a target at test time. To confirm this, we decrease the average number of \emph{green square} distractors per example by about 75\% in the train set and retrain the model. The model performance immediately jumps to 100\%, thereby confirming our hypothesis.

\begin{figure}[t]
	\centering
	\includegraphics[scale=0.55]{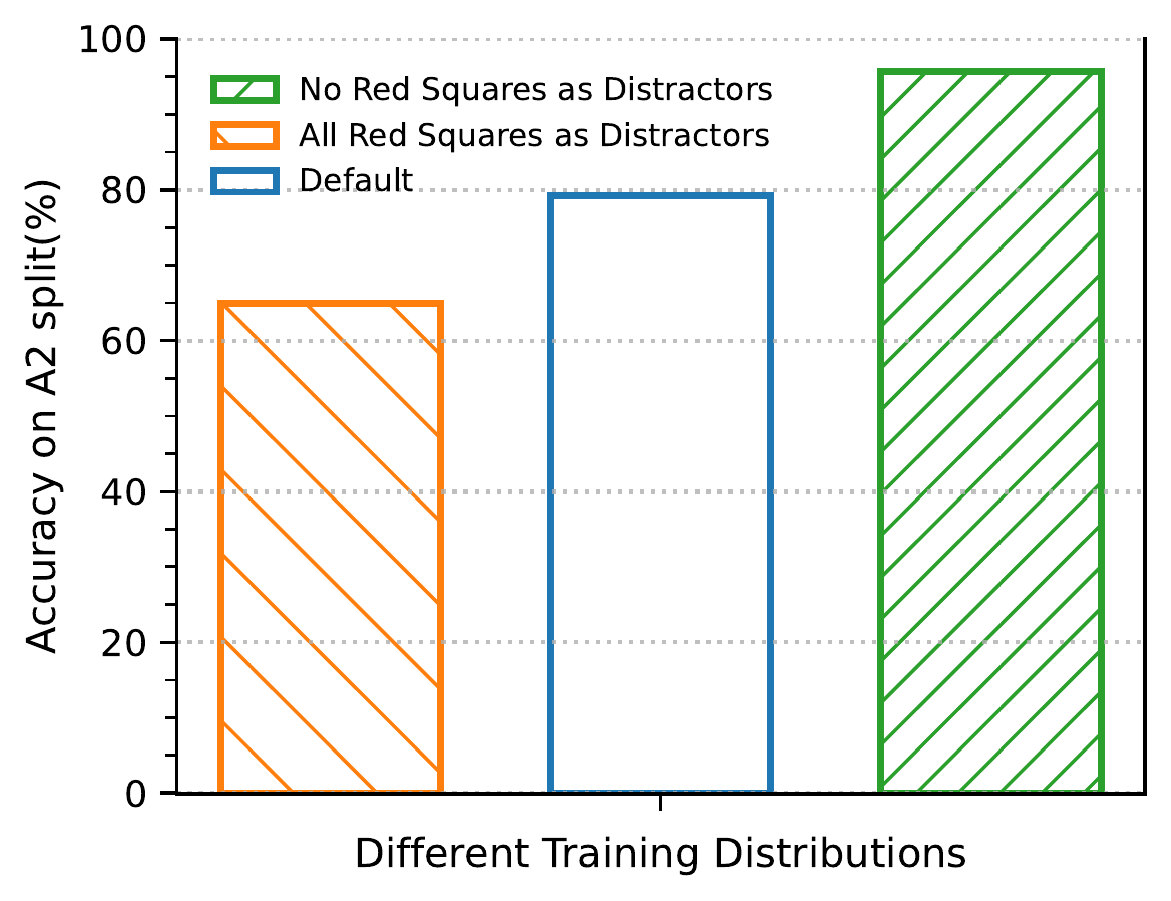}
	\caption{\label{fig:a2_exp} Effect of modifying training distribution on A2 accuracy. The blue bar corresponds to the original training distribution, the orange bar corresponds to a distribution where all red squares occur as distractors, and the green bar corresponds to the training distribution where no red square occurs as a distractor.}
\end{figure}

Similarly, ReaSCAN also has a test split (A2), which is analogous to the A1 split in Referring Expressions. We observe that the Transformer model performs comparatively worse on A2 than other similar splits on ReaSCAN. Considering the above-mentioned result, we believe that the ReaSCAN A2 split is also suffers from the same issue. Hence, we modify the training distribution to vary the number of examples where ``red square'' object occurs as a distractor. We show our results in Figure \ref{fig:a2_exp}. Note that the size of train set in this experiment is 200K examples which is less than half of the full ReaSCAN trainset. Even with less number of training samples, the model trained on the modified training distribution (No examples contain Red Squares as distractors) outperforms the model trained on full ReaSCAN, on the A2 split. This validates our hypothesis about the overfitting issue in transformers.

\begin{figure*}[t]
	\centering
	\includegraphics[scale=0.11]{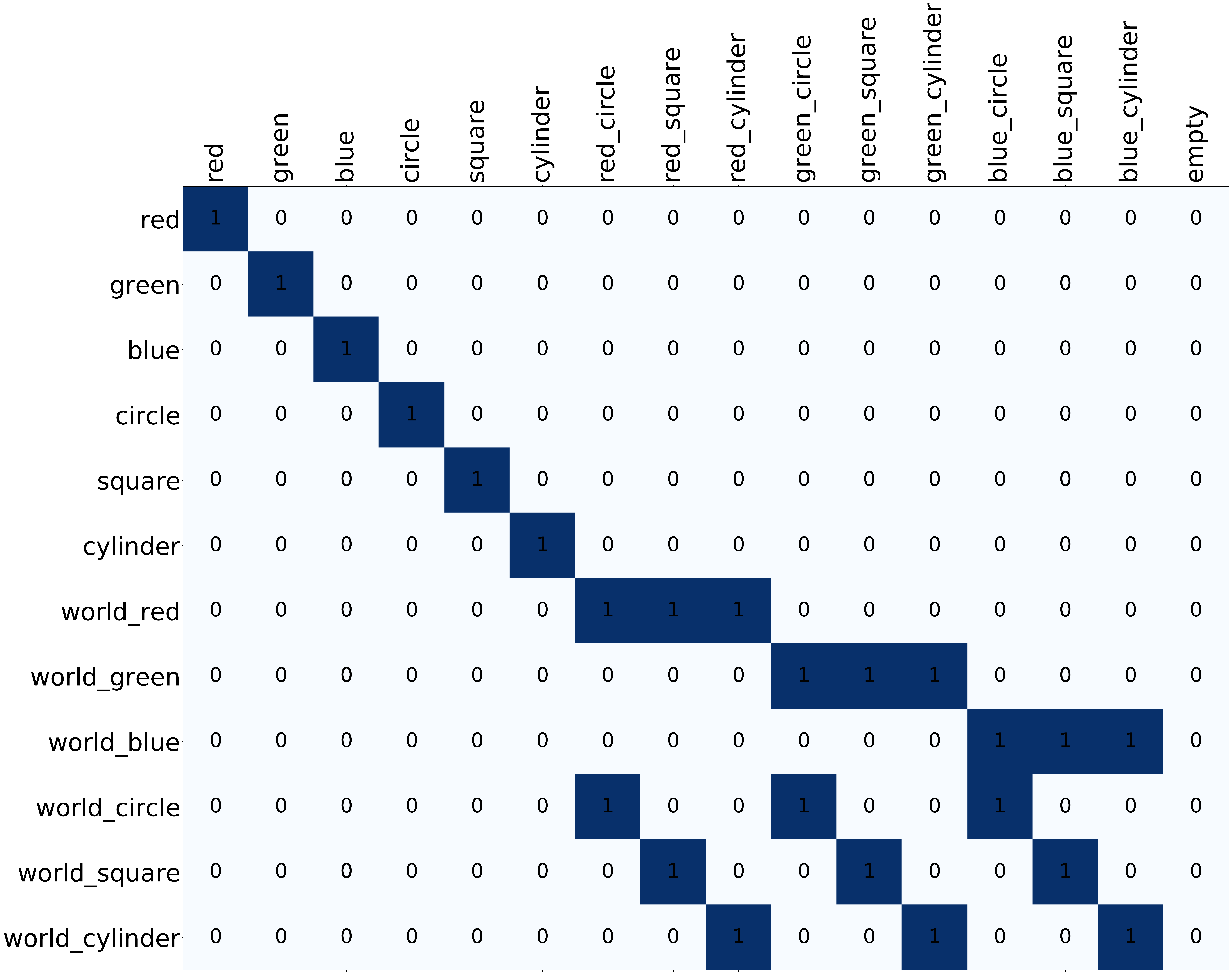}
	\caption{\label{fig:emb_mx} Embedding matrix for \texttt{two-attr} variant. The columns correspond to different tokens in the vocabulary and the rows correspond to what each position in the embedding represents. Along columns, labels like \emph{blue}, and \emph{square} correspond to the command tokens, labels like \emph{red\_circle} correspond to the grid world tokens, where the object has ``red'' color attribute, and ``circle'' shape attribute, and the label \emph{empty} corresponds to the grid world token where there is no object in the grid cell. Along rows, labels like \emph{green} corresponds to the ``green'' color attribute for command tokens, and labels like \emph{world\_green} corresponds to ``green'' color attribute for grid world tokens.}
\end{figure*}

\begin{figure*}[t]
	\centering
	\includegraphics[scale=0.15]{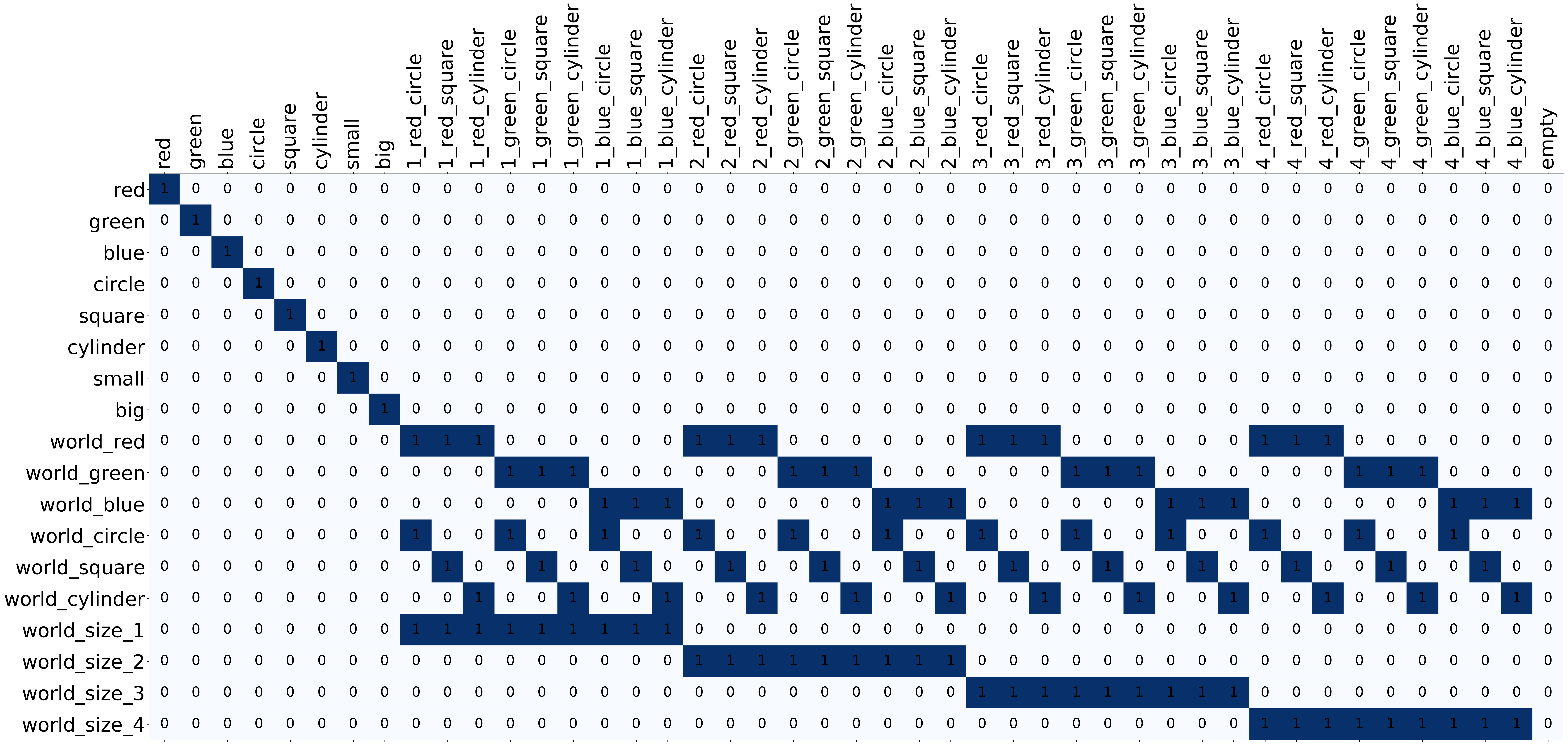}
	\caption{\label{fig:emb_mx_three} Embedding matrix for \texttt{three-attr} variant. The columns correspond to different tokens in the vocabulary and the rows correspond to what each position in the embedding represents. Along columns, labels like \emph{blue}, \emph{square}, and \emph{big} correspond to the command tokens, labels like \emph{1\_red\_circle} correspond to the grid world tokens, where the object has ``1'' size attribute, ``red'' color attribute, and ``circle'' shape attribute, and the label \emph{empty} corresponds to the grid world token where there is no object in the grid cell. Along rows, labels like \emph{green}, \emph{big} correspond to the ``green'' color attribute, and ``big'' size attribute respectively for command tokens, and labels like \emph{world\_green}, \emph{world\_size\_2} corresponds to ``green'' color attribute, and ``2'' size attribute respectively for grid world tokens.}
\end{figure*}

\begin{figure*}[t]
	\centering
	\includegraphics[scale=0.15]{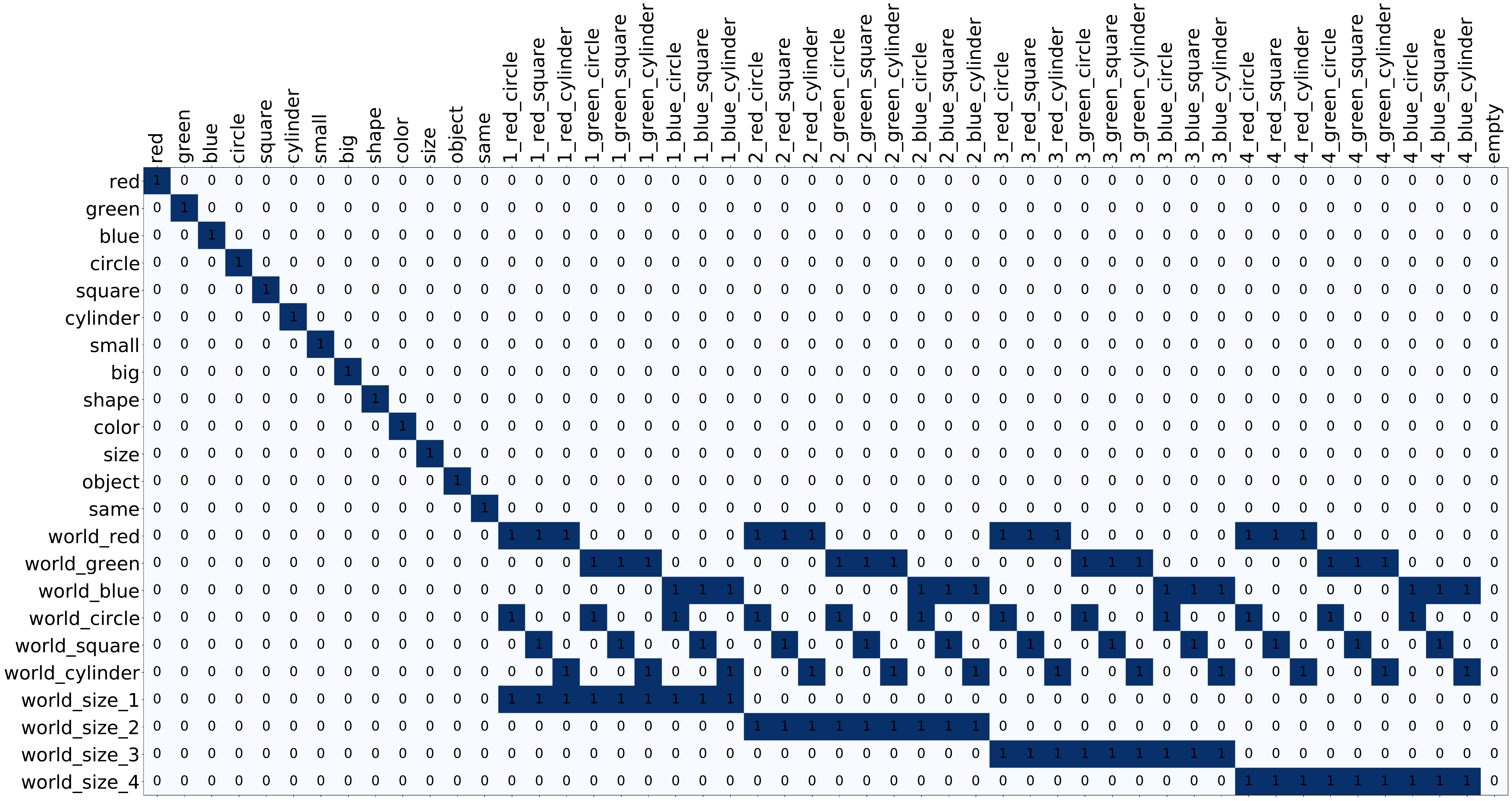}
	\caption{\label{fig:emb_mx_three_rel} Embedding matrix for \texttt{three-attr-rel} variant. Rows and columns have the same meaning as described in Figure \ref{fig:emb_mx_three}.}
\end{figure*}

\begin{figure*}[t]%
	\centering
	\subfloat[\centering Learned Model]{{\includegraphics[scale = 0.15]{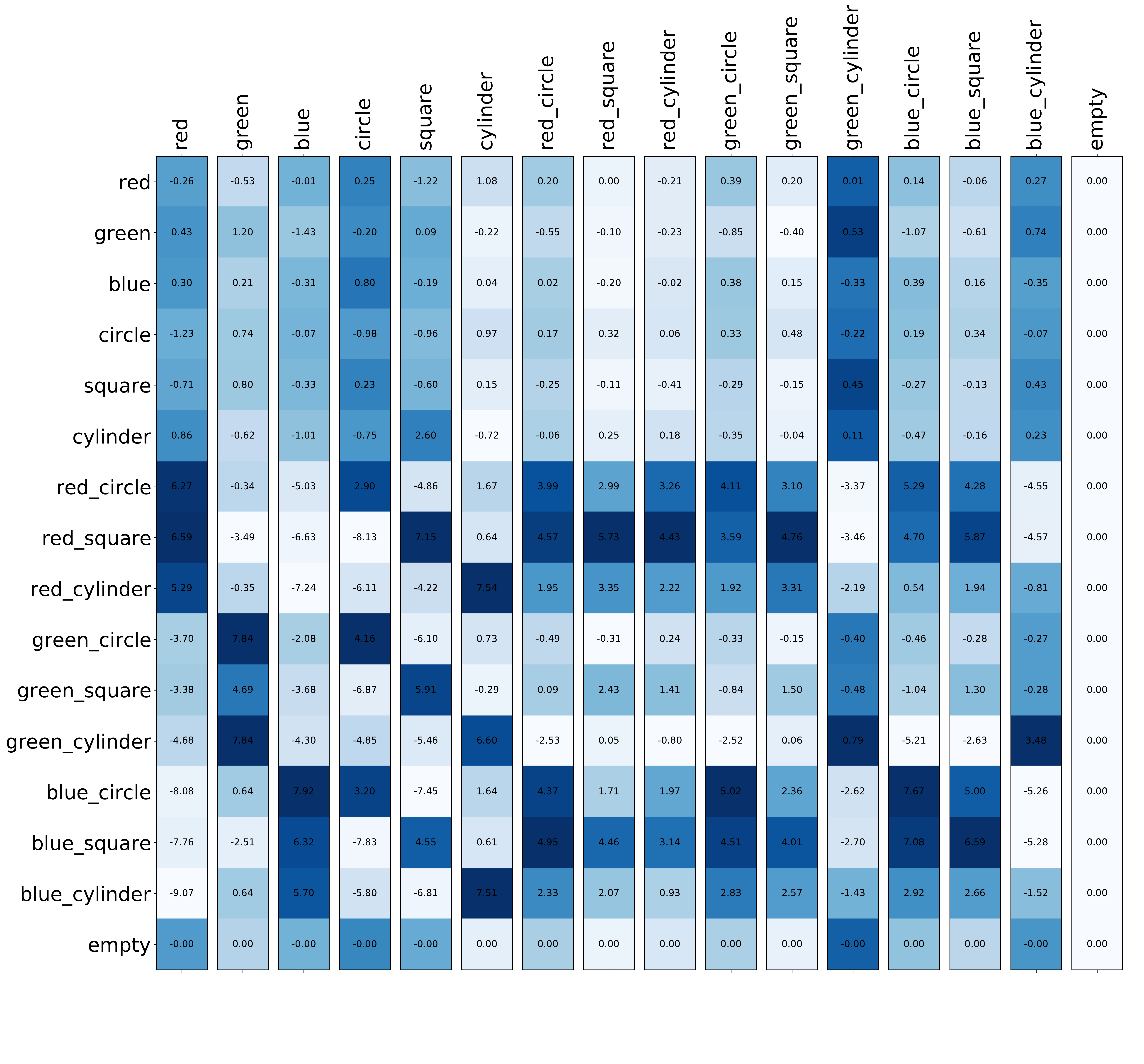} }}%
	%	\qquad
	\subfloat[\centering Our Construction]{{\includegraphics[scale = 0.15]{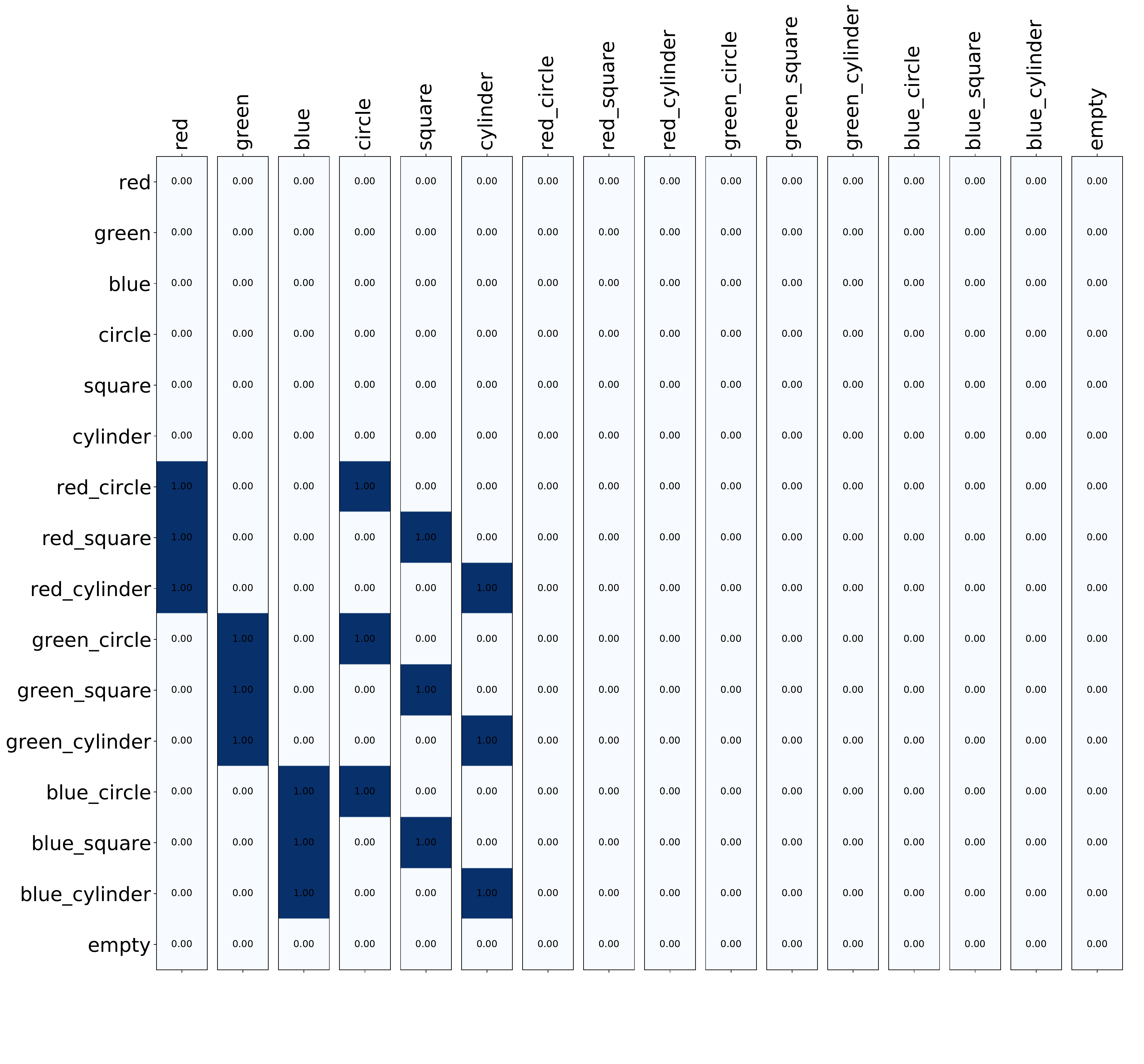} }}%
	
	\caption{$\mathbf{M}_{\mathrm{Learned}}$ (left side) for the attention-only transformer with a single layer and single attention head trained on the \texttt{two-attr} variant and our $\mathbf{M}_{\mathrm{Construct}}$ matrix (right side) for the \texttt{two-attr} variant. Labels like \emph{blue} and \emph{square} correspond to the command tokens, labels like \emph{red\_circle} correspond to the grid world tokens, where the object has ``red'' color attribute and ``circle'' shape attribute, and the label \emph{empty} corresponds to the grid world token where there is no object present in the grid cell. See section \ref{sec:interpret} for the exact formulation of $\mathbf{M}$ matrix.}% 
	\label{fig:qk_matrices_full}%
\end{figure*}

\begin{figure*}[t]%
	\centering
	\subfloat[\centering Learned Model]{{\includegraphics[scale = 0.22]{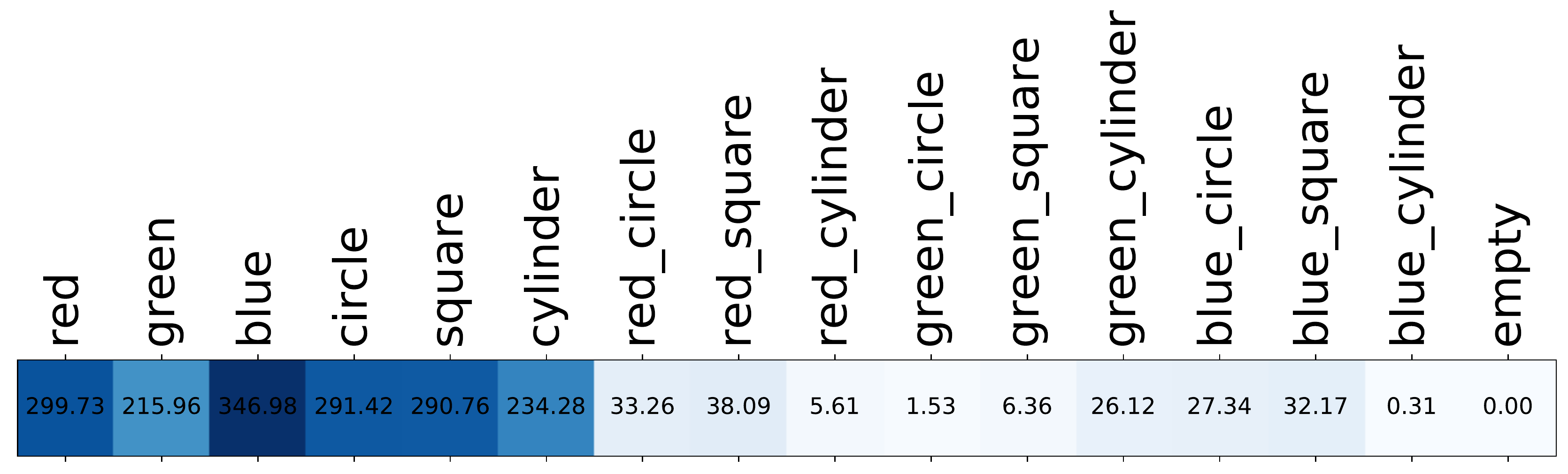} }}%
	%	\qquad
	\subfloat[\centering Our Construction]{{\includegraphics[scale = 0.22]{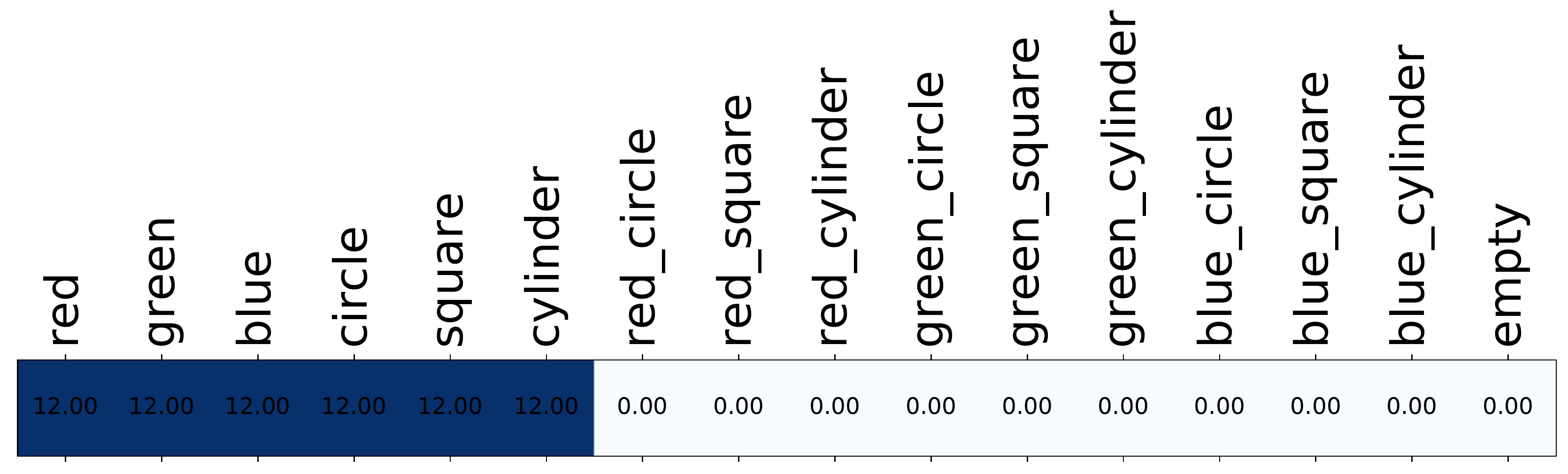} }}%
	
	\caption{$\mathbf{s}_{\mathrm{Learned}}$ (left side) for the attention-only transformer with a single layer and single attention head trained on the \texttt{two-attr} variant and our $\mathbf{s}_{Construct}$ matrix (right side) for the \texttt{two-attr} variant. See section \ref{sec:interpret} for the exact formulation of $\mathbf{s}$. In this particular run, $\mathbf{s}_{\mathrm{Learned}}$ contains positive scalars for command tokens. Some training runs also converged to negative values in $\mathbf{s}_{\mathrm{Learned}}$ for command tokens. In that case, the interpretation of $\mathbf{M}_{\mathrm{Learned}}$ changes accordingly, while the fundamental idea remains the same.}%
	\label{fig:s_full}%
\end{figure*}

\begin{figure*}[t]
	\centering
	\includegraphics[scale=0.30]{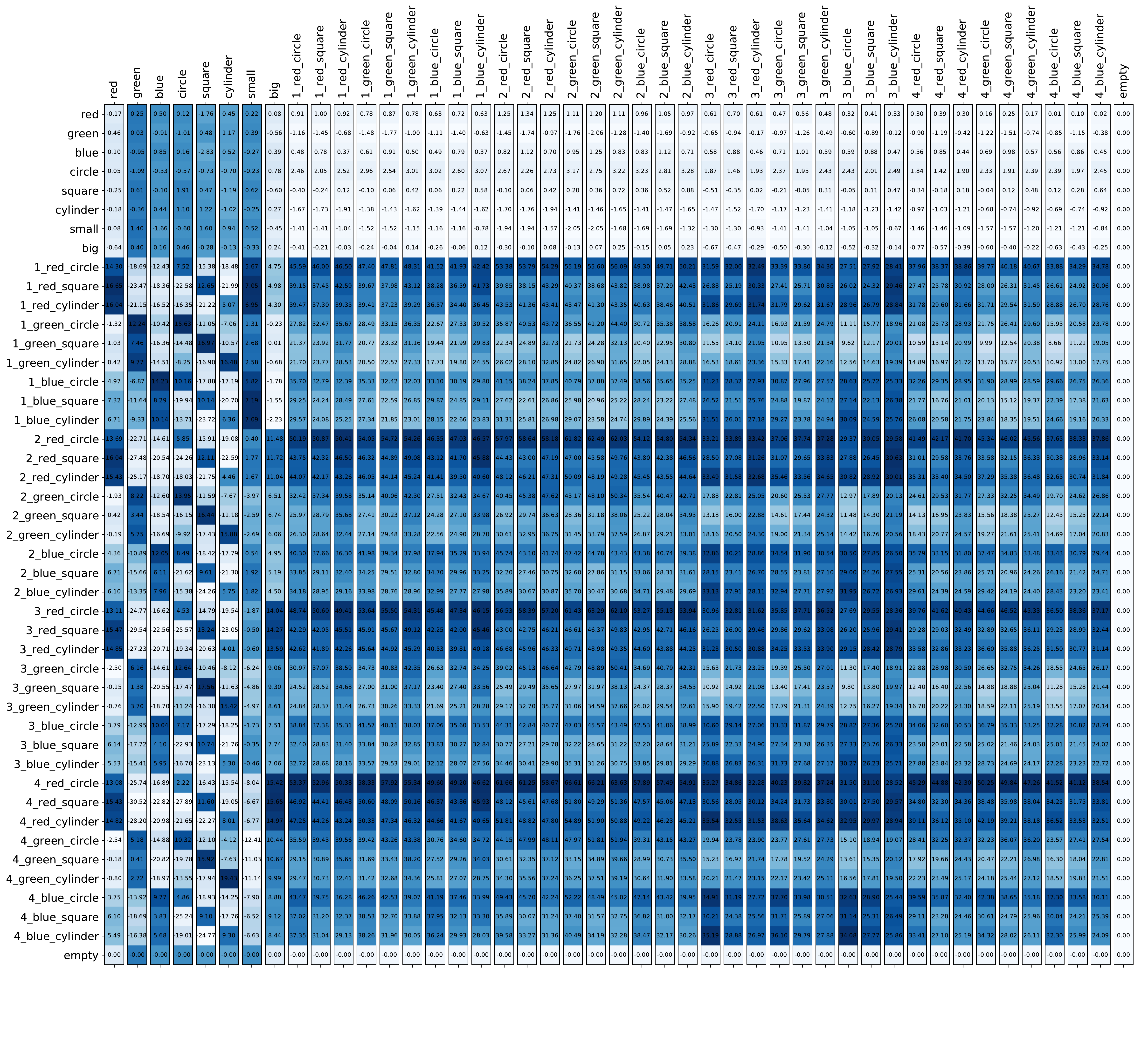}
	\caption{\label{fig:three_attr_m_l} $\mathbf{M}_{\mathrm{Learned}}$ for the attention-only transformer with a single layer and single attention head trained on the \texttt{three-attr} variant. Notice the matching pattern between $\mathbf{M}_{\mathrm{Learned}}$ and $\mathbf{M}_{\mathrm{Construct}}$ (shown in Figure \ref{fig:three_attr_m_c}). We can observe that the dot product between the key of an attribute and queries of grid world tokens with the corresponding attribute has higher values (darker grid cells). For example, in the column of ``square'' command token, the darker grid cells correspond to only those grid world objects which have the ``square'' shape attribute. Similarly, in the column for the ``big'' command token, grid world objects like \emph{4\_blue\_square} with larger size have higher values as compared to objects like \emph{1\_blue\_square} with smaller size.}
\end{figure*}

\begin{figure*}[t]
	\centering
	\includegraphics[scale=0.30]{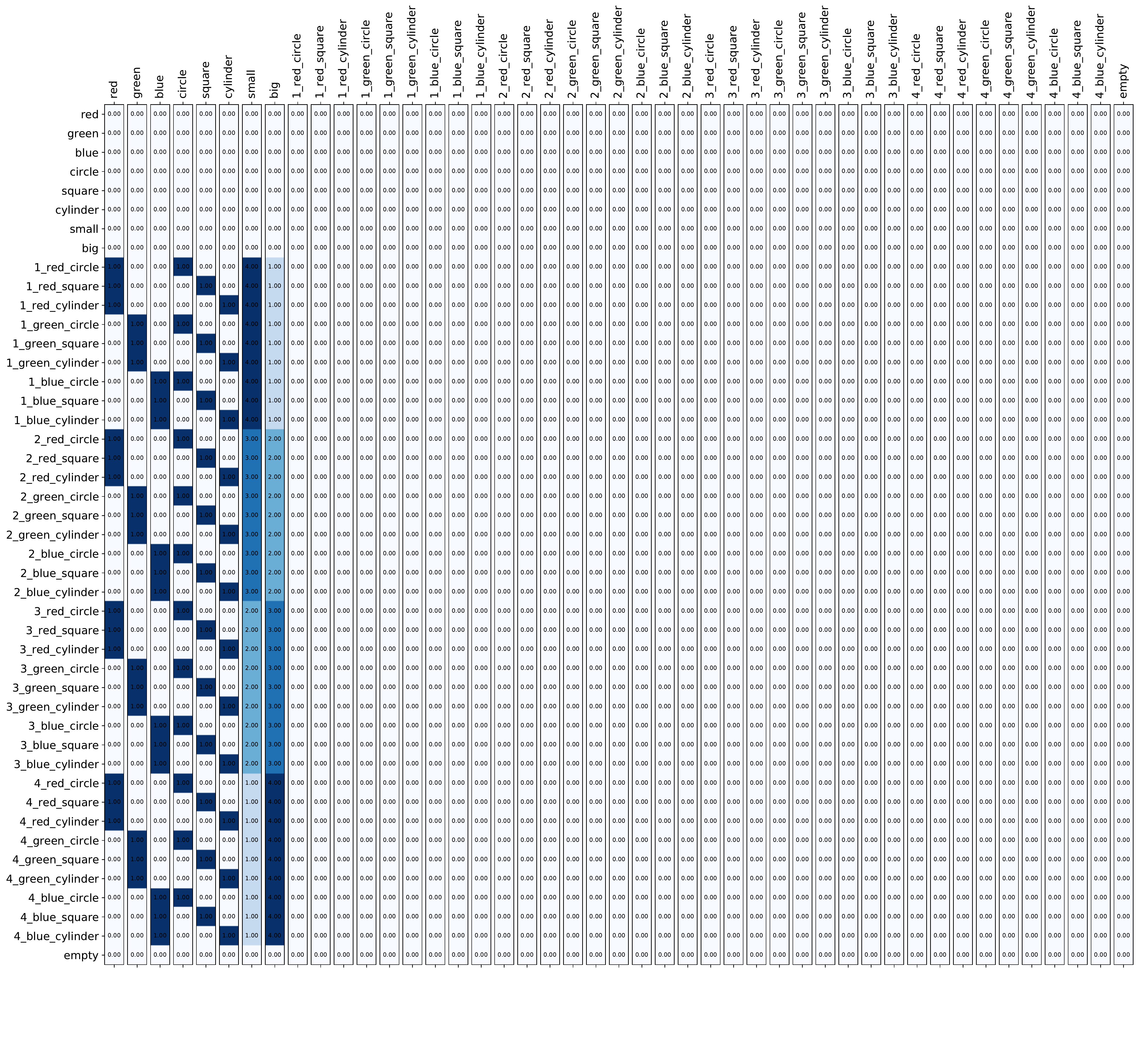}
	\caption{\label{fig:three_attr_m_c} Our $\mathbf{M}_{\mathrm{Construct}}$ matrix for \texttt{three-attr} variant.}
\end{figure*}

\begin{figure*}[htbp]
	\centering
	\includegraphics[scale=0.20]{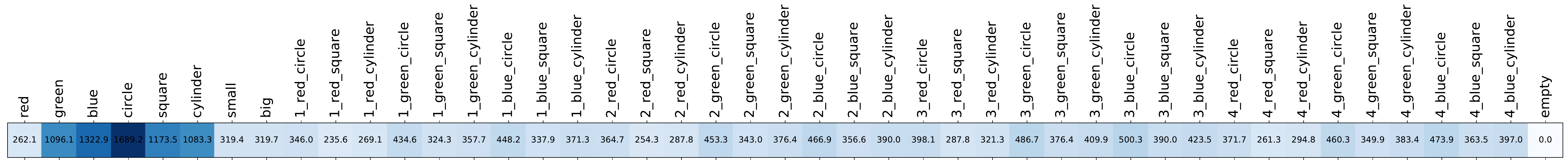}
	\caption{\label{fig:three_attr_s_l} $\mathbf{s}_{\mathrm{Learned}}$ for the attention-only transformer with a single layer and single attention head trained on the \texttt{three-attr} variant. In this particular run, $\mathbf{s}_{\mathrm{Learned}}$ contains positive scalars for command tokens. Some training runs also converged to negative values in $\mathbf{s}_{\mathrm{Learned}}$ for command tokens. In that case, the interpretation of $\mathbf{M}_{\mathrm{Learned}}$ changes accordingly, while the fundamental idea remains the same.}
\end{figure*}

\begin{figure*}[t]
	\centering
	\includegraphics[scale=0.20]{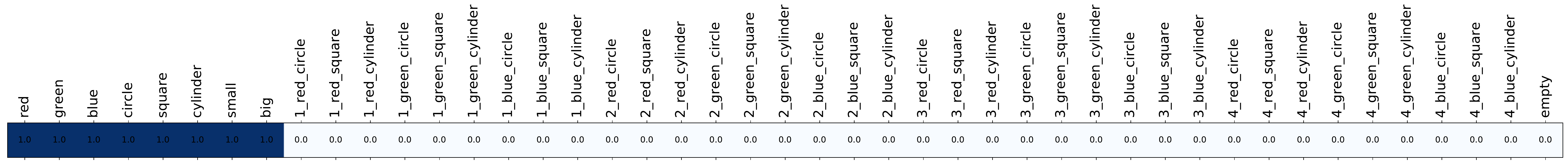}
	\caption{\label{fig:three_attr_s_c} Our $\mathbf{s}_{\mathrm{Construct}}$ for \texttt{three-attr} variant.}
\end{figure*}

\end{document}